\documentclass[11pt]{article}

\usepackage[preprint]{acl}
\usepackage{times}
\usepackage{latexsym}
\usepackage{booktabs}
\usepackage{multirow}
\usepackage{parskip}
\usepackage[T1]{fontenc}

\usepackage[utf8]{inputenc}

\usepackage{microtype}

\usepackage{inconsolata}

\usepackage{graphicx}
\usepackage[many]{tcolorbox}
\tcbuselibrary{breakable}
\usepackage{caption}
\usepackage{fvextra}
\usepackage{fvextra} 
\fvset{
  breaklines=true,
  breaksymbolleft={},  
  breaksymbolright={}  
}
\usepackage{amsfonts}
\DefineVerbatimEnvironment{VerbatimWrap}{Verbatim}{breaklines=true, breakanywhere=true, fontsize=\small}
\usepackage[table]{xcolor}
\definecolor{barbg}{RGB}{253, 224, 224}
\definecolor{barfill}{RGB}{231, 92, 92}
\definecolor{lightbluebg}{RGB}{230,242,255}
\definecolor{lightbluecell}{RGB}{200,225,245}
\definecolor{lightgreencell}{RGB}{210,245,210}
\definecolor{blueframe}{RGB}{41,128,185}
\usepackage{placeins}

\usepackage{hyperref}
\definecolor{citecolor}{HTML}{0071bc}
\hypersetup{
    colorlinks=true,
    linkcolor=red,
    anchorcolor=citecolor,
    citecolor=citecolor
}

\newlength{\barwidth}
\title{
  Dental-TriageBench: Benchmarking Multimodal Reasoning for Hierarchical Dental Triage
  }

\author{
Ziyi He$^{1\ast}$,
Yushi Feng$^{1\ast}$,
Shuangyu Yang$^{2,3}$,
Yinghao Zhu$^{1}$,
Xichen Zhang$^{5}$ \\
\textbf{Tai Pak Chuen Patrick}$^{2,3}$,
\textbf{Lo Hei Yuet}$^{2,3}$,
\textbf{Songying Wu}$^{2,3,4}$,
\textbf{Weifa Yang}$^{2,3\dagger}$,
\textbf{Lequan Yu}$^{1\dagger}$ \\
$^{1}$School of Computing and Data Science, The University of Hong Kong \\
$^{2}$Faculty of Dentistry, The University of Hong Kong \\
$^{3}$The Prince Philip Dental Hospital \\
$^{4}$Li Ka Shing Faculty of Medicine, The University of Hong Kong \\
$^{5}$The Hong Kong University of Science and Technology
}

\begin{document}
\maketitle

\begingroup
\renewcommand\thefootnote{}
\footnotetext{* Equal contribution. $\dagger$ Corresponding authors.}
\endgroup

\begin{abstract}
Dental triage is a safety-critical clinical routing task that requires integrating multimodal clinical information (e.g., patient complaints and radiographic evidence) to determine complete referral plans. We present \textsc{Dental-TriageBench}, the first expert-annotated benchmark for reasoning-driven multimodal dental triage. Built from authentic outpatient workflows, it contains 246 de-identified cases annotated with expert-authored golden reasoning trajectories, together with hierarchical triage labels. We benchmark 19 proprietary, open-source, and medical-domain MLLMs against three junior dentists serving as the human baseline, and find a substantial human--model gap, on fine-grained treatment-level triage. Further analyses show that accurate triage requires both complaint and OPG information, and that model errors concentrate on cases with multiple referral domains, where MLLMs tend to produce overly narrow referral sets and omission-heavy errors. \textsc{Dental-TriageBench} provides a realistic testbed for developing multimodal clinical AI systems that are more clinically grounded, coverage-aware, and safer for downstream care.
\end{abstract}

\section{Introduction}
Dental triage is a safety-critical gatekeeping task at the entry point of care, where clinicians must decide which specialty pathway a patient should enter before definitive chairside tests are available. This decision relies on heterogeneous evidence available at intake. At intake, patients typically present with chief complaints that reflect their perceived symptoms and care needs, but these complaints may not fully localize the underlying problem or reveal concurrent conditions; orthopantomograms (OPGs), by contrast, expose structural findings across the teeth and surrounding maxillofacial region~\cite{turosz2023overview,wang2025multinationaldpr}. Dental triage is therefore inherently multimodal, requiring clinicians to integrate textual complaints with radiographic evidence. It is not merely a perception problem, but a multimodal clinical decision-making task with direct implications for downstream care and patient safety.
\begin{figure}[t]
  \centering
  \includegraphics[width=0.5\textwidth]{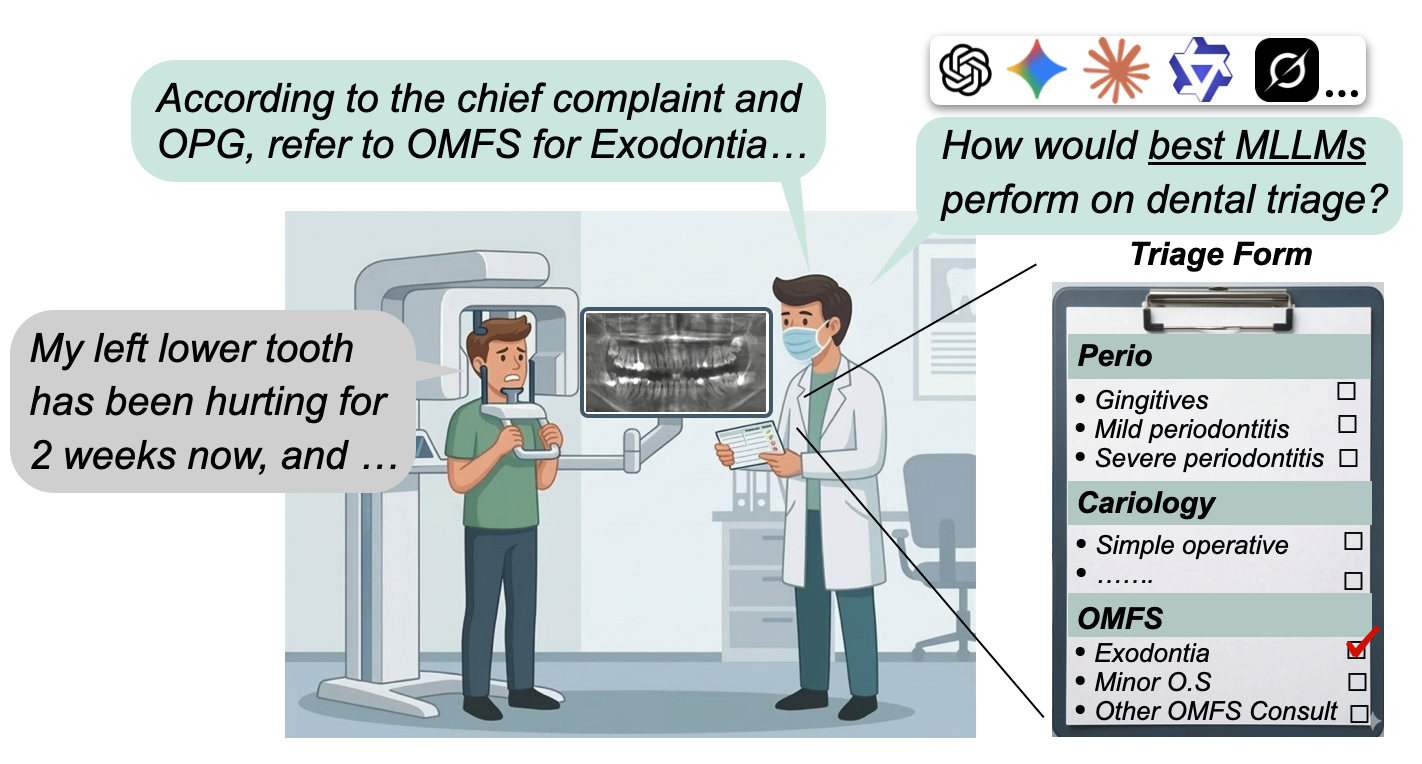}
  \caption{\textbf{Multimodal dental triage.} Given a patient's multimodal information (e.g., chief complaint, orthopantomogram), the model predicts a multi-label specialty referral plan that reflects the patient's treatment needs.}
  \label{fig:teaser}
\end{figure}

\begin{figure*}[t!]
  \centering
  \includegraphics[width=1.0\textwidth]{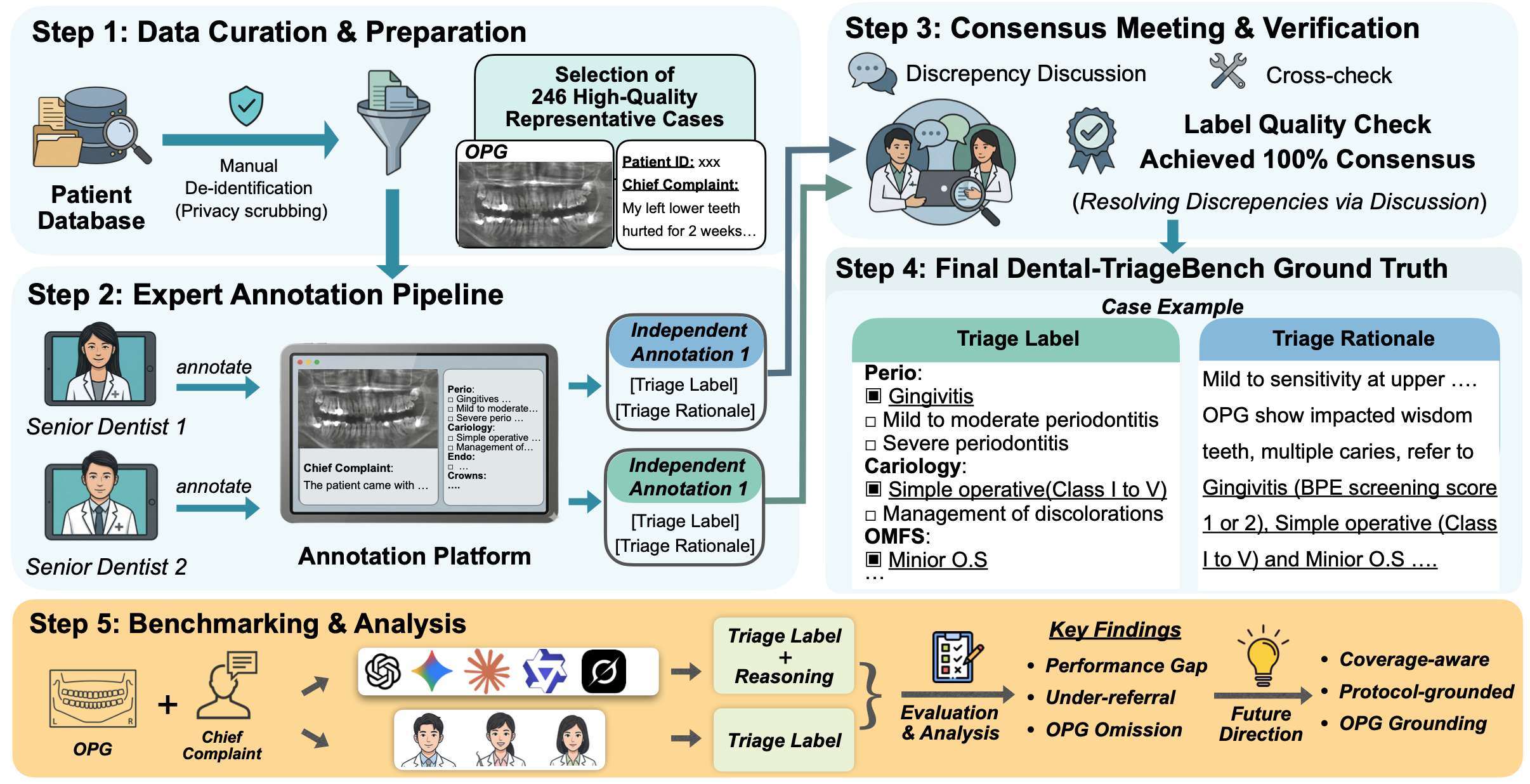}
  \caption{\textbf{Overview of \textsc{Dental-TriageBench}.} Steps 1--4 illustrate the dataset construction pipeline, and Step 5 summarizes the main experimental settings and key findings.}
  \label{fig:Dental_TriageBench}
\end{figure*}

Recent progress in dental AI has substantially improved panoramic radiograph understanding, with benchmarks and models increasingly targeting visual question answering, report generation, and image-grounded dialogue~\cite{turosz2023overview,hao2025mmoral,meng2025dentvlm,huang2025dentvfm,balel2026hybridopgreports}. However, these settings primarily evaluate perception, description, or question answering, and stop short of the core challenge of real dental triage. In practice, triage requires clinicians to jointly reason over patient complaints and radiographic evidence, identify multiple concurrent conditions, and map them to a complete, hierarchical referral plan under clinical protocols. This makes the task fundamentally different from generating plausible image-grounded answers: it is multi-label, protocol-constrained, and safety-sensitive. The distinction is especially important because triage errors are asymmetric: over-referral increases workload, but under-referral leaves clinically relevant problems uncovered and can delay necessary care~\cite{williams2024clinicalacuity,masanneck2024triageperformance,lu2024triageagent}. Existing evaluations therefore do not fully test whether multimodal models can deliver complete and safe referral coverage, rather than merely produce clinically plausible interpretations.


To address this gap, we introduce \textsc{Dental-TriageBench}, the first expert-annotated benchmark for reasoning-driven multimodal dental triage. The benchmark is curated from authentic outpatient workflows where each case in the benchmark consists of an OPG and a patient chief complaint. Every case is annotated by two senior dentists with expert-authored golden reasoning trajectories, together with hierarchical multi-label triage labels spanning 8 coarse-grained domains and 22 fine-grained categories. We formulate dental triage as a hierarchical multimodal multi-label prediction task with rationale generation, and benchmark 19 proprietary, open-source, and medical-domain MLLMs under zero-shot prompting. Our results show that current models remain substantially below human performance and are far from reliable deployment in this safety-critical setting. Even the strongest proprietary models trail junior dentists by a clear margin. We further find a pronounced gap between broad specialty-level routing and clinically precise treatment-level triage: while the best model reaches 0.302 Macro-F1 on the 22 fine-grained triage labels, its performance improves to 0.493 after aggregation to the 8 coarse-grained triage domains. This suggests that current MLLMs are better at identifying the general specialty pathway than at producing complete and treatment-specific referral plans. 

Beyond aggregate scores, we use \textsc{Dental-TriageBench} to study how and why current MLLMs fail on realistic multimodal triage. We find that errors concentrate on cases with multiple referral domains, where models tend to predict referral sets that are systematically narrower than both ground truth and human annotations. To better characterize these failures, we develop a clinician-grounded taxonomy covering radiographic misinterpretation, evidence omission, protocol misalignment, hallucination, and complaint neglect, and scale this analysis with an LLM judge validated against expert annotations. Together, these results suggest that the main bottleneck is not generating plausible dental language, but making complete, protocol-aligned, and coverage-aware triage decisions under realistic multimodal uncertainty.

In summary, our contributions are as follows:
\begin{itemize}
    \item \textbf{We introduce the first benchmark for reasoning-driven multimodal dental triage.} \textsc{Dental-TriageBench} pairs OPGs and chief complaints with expert-annotated hierarchical triage labels and, crucially, expert-authored golden reasoning trajectories that capture clinically grounded referral logic.

    \item \textbf{We provide a rigorous evaluation of current MLLMs against human dentists.} We benchmark 19 proprietary, open-source, and medical-domain MLLMs on this task, and further evaluate three junior dentists on the same benchmark as a human baseline, showing that current models remain substantially below human performance, especially on fine-grained treatment-level triage.

    \item \textbf{We derive actionable insights for safer multimodal clinical AI.} Our analyses show that dental triage requires genuine cross-modal reasoning, and that current MLLMs systematically under-cover required referrals in complex multi-domain cases, highlighting the need for future systems that are more coverage-aware, protocol-aligned, and clinically safe.
\end{itemize}

\section{Related Work}

\subsection{Medical MLLMs for Clinical Evaluation}
Recent work on medical LLMs has moved beyond exam-style QA toward more realistic clinical evaluation. Benchmarks such as ClinicBench, CliMedBench, and LLMEval-Med assess LLMs on authentic clinical scenarios and physician-validated tasks, while triage-specific studies examine emergency acuity assessment and decision support under real or benchmarked triage settings~\cite{liu2024clinicbench,ouyang2024climedbench,zhang2025llmevalmed,williams2024clinicalacuity,masanneck2024triageperformance,lu2024triageagent}. In parallel, medical MLLMs and multimodal benchmarks have expanded image-grounded dialogue, visual question answering, report generation, and clinical visual reasoning across diverse medical modalities~\cite{li2023llavamed,moor2023medflamingo,zhang2024biomedgpt,tu2023generalistbiomedical,chen2024chexagent,gmai2024mmbench,zhou2025drvdbench}. These efforts are closely related to our setting, but they remain predominantly text-centric, diagnosis-centric, or organized around VQA and report generation rather than hierarchical multi-label referral planning with clinician-aligned failure analysis~\cite{feng2026pass,vqa}.

\subsection{Dental AI on panoramic radiographs}
In dentistry, AI on panoramic radiographs has historically centered on task-specific computer vision problems such as tooth detection and numbering, restoration or state classification, and multi-disease diagnosis~\cite{estai2022toothnumbering,zhu2023multipledentaldiseases}. Recent reviews confirm rapid progress in dentomaxillofacial imaging, while also noting that most existing studies remain narrowly task-specific and only weakly connected to real clinical workflows~\cite{turosz2023overview}. More recent dental multimodal efforts begin to move toward broader foundation and instruction-following settings, including MMOral and OralGPT for panoramic-X-ray question answering, dialogue, and reporting, DentVLM for comprehensive multimodal dental diagnosis, DentVFM for generalist dental representation learning, OPGAgent for auditable OPG interpretation, OMNI-Dent for clinically guided explainable diagnosis, and recent hybrid LLM pipelines for panoramic report generation~\cite{hao2025mmoral,meng2025dentvlm,huang2025dentvfm,jang2026omnident,balel2026hybridopgreports}. However, existing dental benchmarks still largely formulate image interpretation as diagnosis, reporting, or question answering; they do not directly evaluate the outpatient triage problem in which complaints and OPG findings must be jointly mapped to complete, hierarchical, multi-domain referral decisions.

\section{Dental-TriageBench}
We introduce \textsc{Dental-TriageBench}, a clinically grounded multimodal benchmark for dental triage. Each case consists of a patient’s chief complaint and an orthopantomogram (OPG), with expert-annotated multi-label triage decisions and supporting clinical rationale.

\subsection{Data Curation}
All cases were retrospectively collected from historical clinical records of dental school clinics. To protect patient privacy, all textual notes and OPG images were manually de-identified before annotation, with personal identifiers removed from both modalities. After excluding low-quality or clinically inconclusive samples, the final benchmark contains 246 patient cases.
Each case was annotated by two licensed senior dentists. For every chief complaint--OPG pair, the annotators provided:  (1) \textbf{structured triage labels}, and  (2) \textbf{clinical rationale} explaining how the multimodal evidence supports the triage decision. 

The triage taxonomy is clinically hierarchical, consisting of \textbf{8 coarse-grained triage domains} and \textbf{22 fine-grained triage labels} that cover major referral destinations and treatment types in routine dental triage. The full taxonomy is provided in Appendix ~\ref{sec:appendix_triage_taxonomy}. The two dentists first reviewed all cases independently and then resolved disagreements through discussion to obtain the final ground truth.

\begin{figure}[t]
  \centering
  \includegraphics[width=0.5\textwidth]{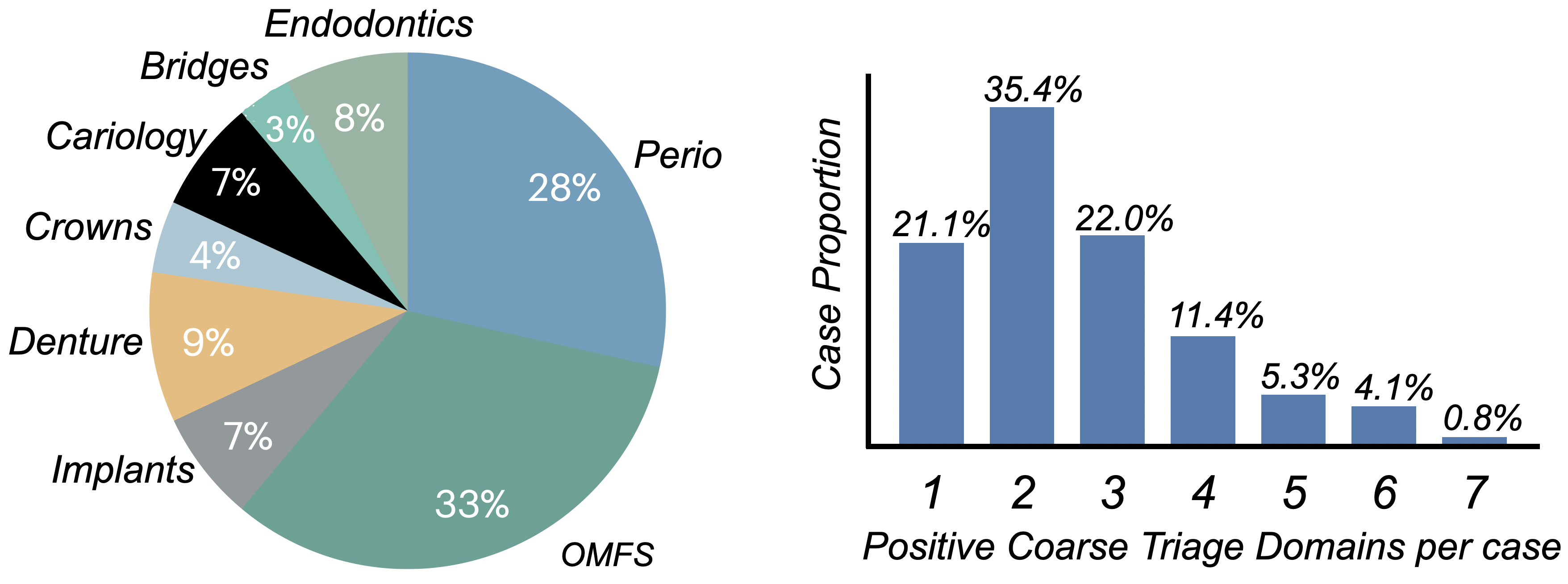}
  \caption{Data statistics of \textsc{Dental-TriageBench}.}
  \label{fig:benchmark_stat}
\end{figure}

\subsection{Benchmark Characteristics}
As shown in the bar chart of Figure~\ref{fig:benchmark_stat}, the label distribution is strongly long-tailed. Common triage domains such as OMFS and Periodontology account for a large fraction of positive labels, whereas domains such as Bridges and Crowns appear much less frequently. We retain this naturally imbalanced distribution because it reflects authentic outpatient triage rather than a benchmark constructed for class balance. As a result, the task is more clinically realistic and more demanding: models must not only identify common referral patterns, but also preserve coverage over less frequent yet clinically relevant triage categories.

Second, \textsc{Dental-TriageBench} preserves the complexity of real clinical triage rather than artificially balancing cases. First, the benchmark is inherently \textbf{multi-label}: At the coarse-grained domain level, each case is associated with \textbf{2.60 positive triage domains} on average, as shown in the pie chart in Figure~\ref{fig:benchmark_stat}, only 21.1\% of patients require referral to a single domain, the remaining \textbf{78.9\%} involve multiple co-occurring conditions and multi-domain referral needs. This combination of long-tailed label prevalence and multi-label complexity makes \textsc{Dental-TriageBench} a challenging testbed for current MLLMs.

\subsection{Task Definition}
We formulate dental triage as a \textbf{multimodal multi-label classification} task with rationale generation. Given a patient case $\mathcal{X} = (\mathcal{X}_{img}, \mathcal{X}_{text}), $where $\mathcal{X}_{img}$ is the OPG and $\mathcal{X}_{text}$ is the chief complaint, the model predicts $\mathcal{Y} = (\mathbf{y}, \mathcal{R}).$

Here, $\mathbf{y} \in \{0,1\}^{|\mathcal{L}|}$ is a multi-hot label vector over the 22 fine-grained triage labels, and $\mathcal{R}$ is a natural-language rationale that explains the predicted triage decision. Because patients frequently present with multiple concurrent conditions, this task is inherently multi-label rather than single-label.

\subsection{Evaluation}
We evaluate structured triage prediction using \textit{Macro-F1}, \textit{Macro-Recall}, \textit{Micro-F1}, and \textit{Exact Match Ratio (EMR)}. These metrics capture complementary aspects of performance under severe class imbalance and multi-label prediction: class-balanced quality, overall label recovery, and exact case-level correctness. Additional details are provided in Appendix \ref{subsec:appendix_metrics}.

\begin{table*}[htbp!] 
\centering 
\footnotesize 
\setlength{\tabcolsep}{3.5pt} 
\resizebox{\textwidth}{!}{%
\begin{tabular}{l|cccc|cccc} 
\toprule[1.5pt] 
\multirow{2}{*}{\textbf{Model}} & \multicolumn{4}{c|}{\textbf{Fine-grained (22 labels)}} & \multicolumn{4}{c}{\textbf{Coarse-grained (8 domains)}} \\ \cmidrule(lr){2-5}\cmidrule(lr){6-9} 
& \textbf{Macro-F1} & \textbf{Macro-Rec} & \textbf{Micro-F1} & \textbf{Exact-Match} & \textbf{Macro-F1} & \textbf{Macro-Rec} & \textbf{Micro-F1} & \textbf{Exact-Match} \\ 

\multicolumn{9}{l}{\cellcolor{lightbluecell}\textit{\textbf{Proprietary MLLMs}}} \\ 
Claude-Sonnet-4-5-20250929 & 0.269 & \underline{0.275} & 0.350 & 0.024 & \underline{0.442} & \underline{0.466} & 0.565 & 0.130 \\ 
Gemini-3-Flash & \textbf{0.302} & \textbf{0.310} & \textbf{0.459} & \underline{0.049} & \underline{0.488} & \textbf{0.492} & \textbf{0.656} & \underline{0.220} \\ 
Gemini-3-Pro & \underline{0.294} & 0.262 & \underline{0.447} & \textbf{0.118} & \textbf{0.493} & 0.444 & \underline{0.641} & \textbf{0.224} \\ 
GPT-5.2 & 0.272 & 0.239 & 0.404 & 0.041 & 0.398 & 0.320 & 0.576 & 0.215 \\ 
Grok-4 & 0.232 & 0.187 & 0.329 & 0.094 & 0.392 & 0.317 & 0.529 & 0.187 \\ 
Kimi-K2.5 & 0.249 & 0.204 & 0.348 & 0.053 & 0.393 & 0.311 & 0.537 & 0.187 \\ 
Qwen-VL-Max-Latest & 0.222 & 0.206 & 0.283 & 0.037 & 0.398 & 0.319 & 0.517 & 0.183 \\ 

\hline 
\multicolumn{9}{l}{\cellcolor{lightbluecell}\textit{\textbf{Open-source MLLMs}}} \\ 
GLM-4.1V-9B-Thinking & 0.072 & 0.050 & 0.086 & 0.016 & 0.133 & 0.089 & 0.182 & 0.069 \\ 
InternVL3\_5-8B & 0.088 & 0.056 & 0.093 & 0.016 & 0.176 & 0.104 & 0.196 & 0.045 \\ 
MiMo-VL-7B-RL & \underline{0.125} & \underline{0.093} & \underline{0.175} & 0.020 & \underline{0.266} & \underline{0.184} & \underline{0.386} & \underline{0.146} \\ 
Ministral-3-14B-Instruct-2512 & \textbf{0.189} & \textbf{0.158} & \textbf{0.247} & \textbf{0.029} & \textbf{0.327} & \textbf{0.254} & \textbf{0.453} & \textbf{0.163} \\ 
Qwen2.5-VL-7B-Instruct & 0.101 & 0.077 & 0.148 & \underline{0.024} & 0.191 & 0.135 & 0.301 & 0.102 \\ 
Qwen3-VL-8B-Instruct & 0.089 & 0.064 & 0.127 & \underline{0.024} & 0.208 & 0.131 & 0.281 & 0.106 \\ 
LLaVA-onevision-qwen2-7b & 0.064 & 0.049 & 0.095 & 0.000 & 0.134 & 0.084 & 0.187 & 0.045 \\ 
LLaVA-v1.6-mistral-7b-hf & 0.014 & 0.008 & 0.019 & 0.004 & 0.032 & 0.017 & 0.058 & 0.008 \\ 

\multicolumn{9}{l}{\cellcolor{lightbluecell}\textit{\textbf{Medical MLLMs}}} \\ 
HuatuoGPT-Vision-7B & 0.074 & \underline{0.058} & 0.101 & 0.004 & 0.121 & 0.072 & 0.188 & 0.061 \\ 
Lingshu-7B & \underline{0.076} & 0.056 & \underline{0.115} & \underline{0.012} & \underline{0.172} & \underline{0.108} & \underline{0.277} & \underline{0.094} \\ 
MedGemma-27b-it & \textbf{0.121} & \textbf{0.094} & \textbf{0.173} & \textbf{0.029} & \textbf{0.266} & \textbf{0.194} & \textbf{0.357} & \textbf{0.130} \\ 
MedGemma-1.5-4b-it & 0.059 & 0.036 & 0.068 & 0.008 & 0.116 & 0.068 & 0.140 & 0.041 \\ 
\midrule[1.2pt] 
\rowcolor{lightgreencell}
\rule{0pt}{3pt} \textit{\textbf{Human Baseline}} & \textbf{0.402} & \textbf{0.431} & \textbf{0.525} & \textbf{0.083} & \textbf{0.562} & \textbf{0.613} & \textbf{0.727} & \textbf{0.272} \\ 

\bottomrule[1.5pt] 
\end{tabular}%
}
\caption{Benchmark results on \textsc{Dental-TriageBench} across 22 fine-grained triage labels and 8 aggregated coarse-grained domains. We report Macro-F1, Macro-Recall, Micro-F1, and Exact-Match for 19 proprietary, open-source, and medical MLLMs, together with the human baseline. Within each model domain, the best results are in \textbf{bold} and the second-best results are \underline{underlined}.} 
\label{tab:merged_fine_coarse_tab} 
\end{table*}

\section{Experiments}

\subsection{Experiment Setup}
To evaluate the capability boundaries of current MLLMs on specialized clinical decision-making, we benchmark 19 representative multimodal models, including 7 proprietary models~\cite{anthropic2025claude45,google2026gemini3flashpreview,google2026gemini3propreview,openai2025gpt52,xai2025grok4,kimi2026k25,alibaba2026qwenvlmax}, 8 open-source general-purpose models~\cite{glmv2025glm41vthinking,wang2025internvl35,xiaomi2025mimovl,liu2026ministral3,bai2025qwen25vl,bai2025qwen3vl,li2024llavaonevision,li2024llavanext}, and 4 medical-domain models~\cite{chen2024huatuogptvision,lasa2025lingshu,google2026medgemma15,google2025medgemma27}. All models are evaluated in a zero-shot setting to assess their inherent clinical alignment, and the full task instruction prompt is provided in Table~\ref{fig:task_instruction}.

For each case, models are asked to predict the exact triage decisions in the \textit{22 fine-grained triage label space} in a single run. Our primary evaluation is therefore conducted on this \textit{fine-grained multi-label prediction task}. To additionally assess whether models can still recover broader referral needs, we aggregate the 22 fine-grained label predictions into their corresponding \textit{8 coarse-grained triage domains} (e.g., mapping \textit{Single-unit Crowns} to \textit{Crowns}) and report performance at this higher level as well. To provide a clinically meaningful reference point, we further invited three junior dentists to complete the full benchmark under the same fine-grained label space. We report the average performance of the three dentists as the human baseline in the main paper, and provide the individual scores of each dentist in the Appendix ~\ref{subsec:appendix_human_baseline}.

\subsection{Main Results}
\label{sec:main_results}
We report the overall performance in Table~\ref{tab:merged_fine_coarse_tab}, and we make three observations below.

\paragraph{Current MLLMs remain substantially below human performance on dental triage}
Overall, current MLLMs remain substantially below \textbf{junior dentists}, especially on the clinically more demanding fine-grained triage task. On the \textit{22 fine-grained triage labels}, the best-performing proprietary model, Gemini-3-Flash, achieves a Macro-F1 of 0.302 and a Micro-F1 of 0.459, while the junior dentists average reaches 0.402 and 0.525, respectively. This gap persists across all four metrics, indicating that even the strongest frontier MLLMs still struggle to match human-level performance in precise treatment-level triage. The gap becomes even more notable when considering Macro-Recall, where junior dentists achieve 0.431, suggesting that human raters are better at covering the full set of required referrals in multi-label cases.

\begin{figure}[h!]
  \centering
  \includegraphics[width=1\linewidth]{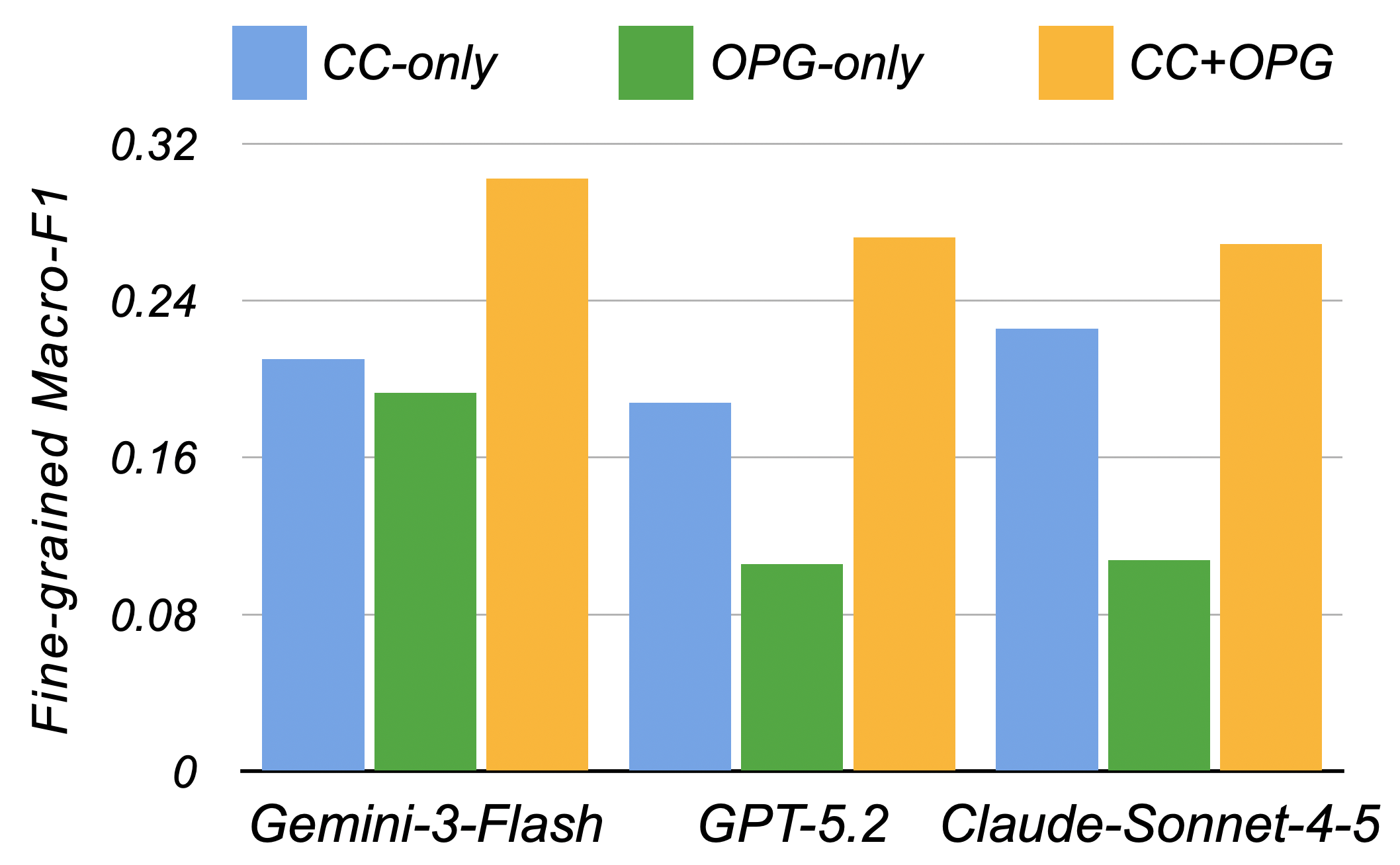}
  \caption{CC-only: only chief complaint; OPG-only: only OPG; CC+OPG: both chief complaint and OPG}
  \label{fig:modality_ablation}
\end{figure}

\paragraph{Proprietary frontier models consistently outperform open-source models}
Gemini-3-Flash yields the strongest overall \textbf{fine-grained} performance, ranking first on Macro-F1, Macro-Recall, and Micro-F1, whereas Gemini-3-Pro achieves the best Exact-Match score (0.118). In contrast, open-source and medical MLLMs lag far behind proprietary ones. The best open-source model, Ministral-3-14B-Instruct-2512, reaches only 0.189 Macro-F1 and 0.247 Micro-F1 on the fine-grained task, while the strongest medical model, MedGemma-27b-it, achieves 0.121 Macro-F1 and 0.173 Micro-F1. Notably, domain-specific medical MLLMs do not outperform strong general-purpose open models, indicating that current medical adaptation alone does not suffice for this highly specialized multimodal dental triage task.

\paragraph{Coarse-grained domain-level routing is more clinically deployable than fine-grained triage prediction}
When predictions are aggregated from the \textit{22 fine-grained triage labels} into \textit{8 coarse-grained triage domains}, all models improve substantially, while the overall ranking pattern remains largely unchanged. For example, Gemini-3-Flash improves from 0.302 to 0.488 in Macro-F1 and from 0.459 to 0.656 in Micro-F1. This suggests that current MLLMs are better suited to broad specialty-level routing than to precise treatment-level triage, making coarse-grained routing a more realistic near-term deployment target. However, even in this setting, the best model still remains clearly below human performance.

\begin{figure}[t!]
  \centering
  \includegraphics[width=1\linewidth]{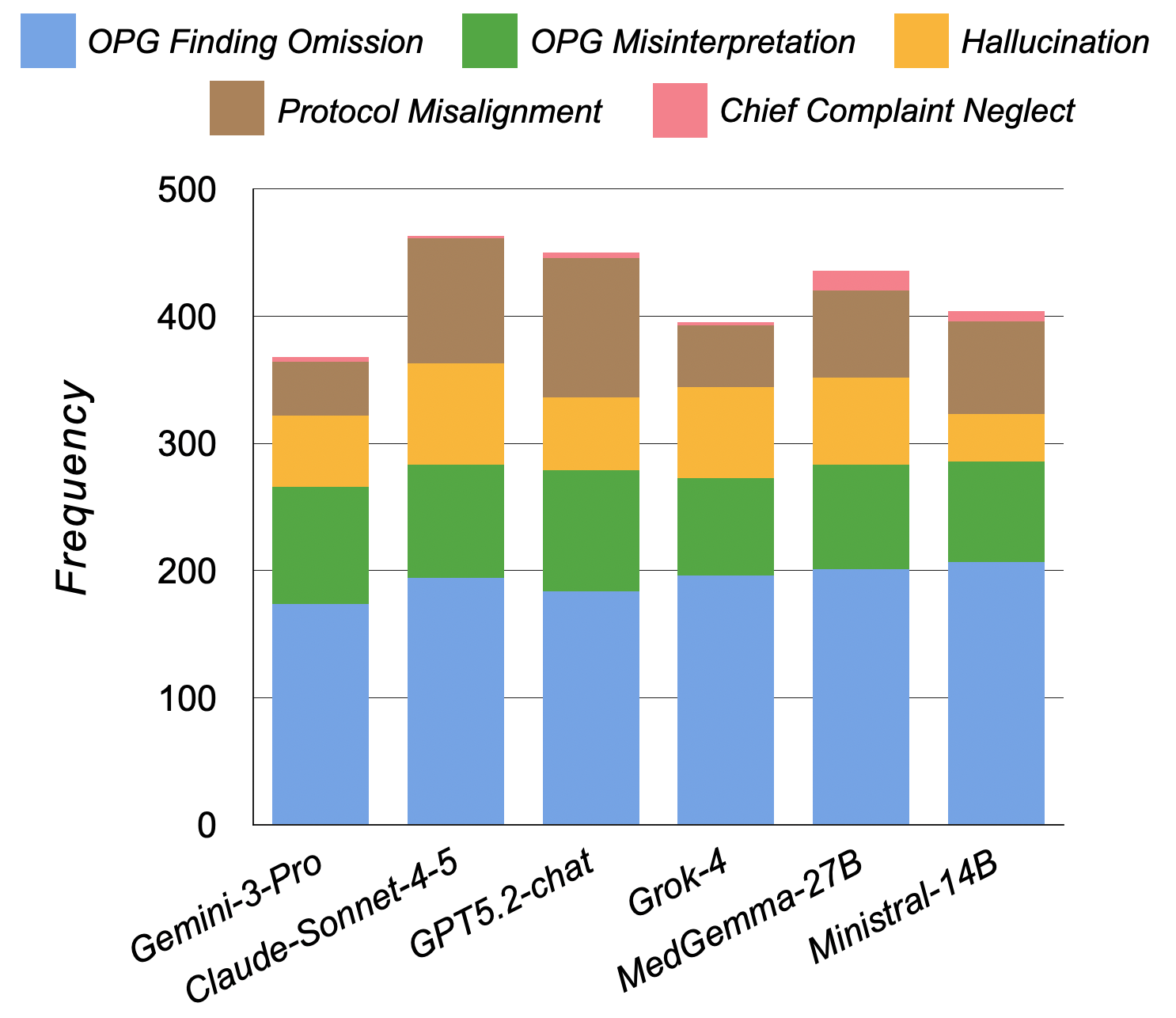}
  \caption{Distribution of five clinically grounded failure dimensions across six representative MLLMs, estimated by Gemini-3-Pro over all benchmark cases.}
  \label{fig:failure_dimension}
\end{figure}

\section{Further Analysis}
\subsection{Multimodality is essential for dental triage}
In this section, we examine the necessity of multimodal input for dental triage. The fundamental premise of \textsc{Dental-TriageBench} is that referral decisions must be made from both what the patient reports and what the radiograph reveals: The chief complaints encode symptom location and care intent, whereas OPGs expose structural and often asymptomatic findings that may independently trigger referral. We therefore evaluate three top-performing MLLMs under chief complaint-only, OPG-only, and multimodal settings. Figure~\ref{fig:modality_ablation} shows that removing either modality degrades performance for all 3 models, i.e., $F1_{\text{Fusion}} > \max(F1_{\text{Text}}, F1_{\text{Image}})$, confirming that neither input source alone is sufficient for reliable triage. \textit{We further provide a representative case in Appendix ~\ref{subsec:appendix_modality_ablation} to illustrate what information is lost in each unimodal setting.}

Crucially, the value of multimodal input becomes even clearer at the case level. As shown in Figure~\ref{tab:case_gain_full}, when measured at the case level, adding the missing modality yields case-level F1 gains greater than 0.1 for 13.0\% of cases for Claude Sonnet 4.5, 17.5\% for Gemini 3 Flash, and 34.1\% for GPT-5.2; gains greater than 0.2 are observed for 11.0\%, 10.2\%, and 27.6\% of cases, respectively. Taken together, these results confirm that our benchmark probes genuine cross-modal evidence integration rather than unimodal shortcut learning.

\begin{table}[h!]
\centering
\small
\setlength{\tabcolsep}{4pt}
\begin{tabular}{lccc}
\toprule
Model & $\Delta$F1 $> 0$ & $\Delta$F1 $> 0.1$ & $\Delta$F1 $> 0.2$ \\
\midrule
Claude Sonnet 4.5 & 17.1\% & 13.0\% & 11.0\% \\
Gemini 3 Flash    & 26.0\% & 17.5\% & 10.2\% \\
GPT-5.2           & 35.4\% & 34.1\% & 27.6\% \\
\bottomrule
\end{tabular}
\caption{Percentage of cases with positive case-level F1 gain after adding the missing modality.}
\label{tab:case_gain_full}
\end{table}

\begin{figure}[t!]
  \centering
  \includegraphics[width=\linewidth]{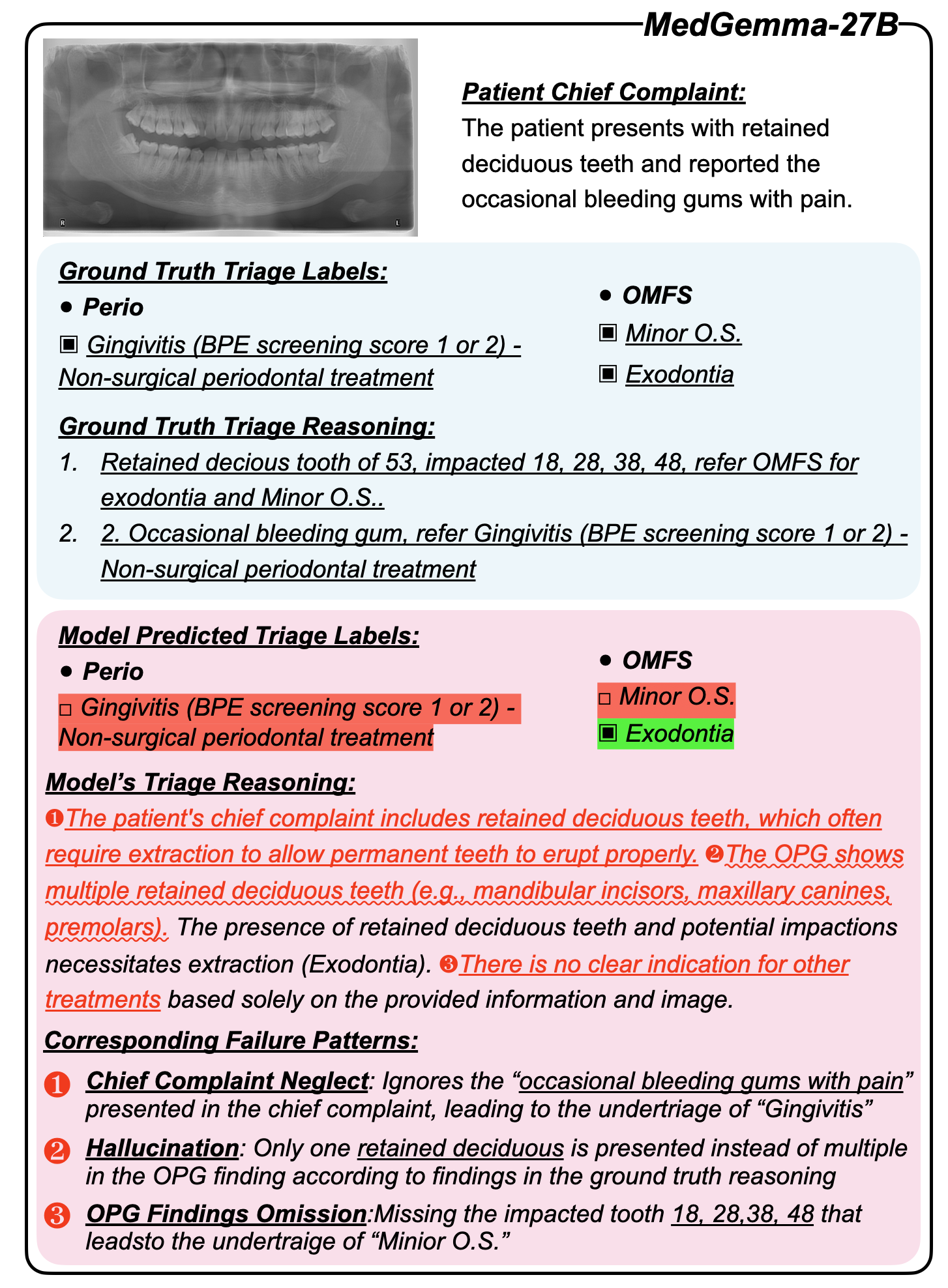}
  \caption{A failure taxonomy case study on MedGemma-27B, with correct triage label on \colorbox{green!65}{green}, and missing triage label on \colorbox{red!65}{red}.}
  \label{fig:failure_patterns_case_study_medgemma}
\end{figure}

\subsection{Clinician-Grounded Failure Taxonomy}
\label{sec: failure_analysis}
As shown in the main results, all evaluated MLLMs underperform human dentists on \textsc{Dental-TriageBench}. However, aggregate metrics such as F1 and Exact Match do not reveal why these models fail or whether their errors follow clinically meaningful patterns. We therefore conduct a finer-grained failure analysis of model-generated rationales. Senior dentists first reviewed a subset of outputs and identified five recurring failure modes: \textit{(1) OPG Misinterpretation}, \textit{(2) OPG Finding Omission}, \textit{(3) Protocol Misalignment}, \textit{(4) Hallucination}, and \textit{(5) Chief Complaint Neglect}. Detailed definitions are provided in Appendix~\ref{subsec:appendix_dimension_definition}.

\begin{figure}[t!]
  \centering
  \includegraphics[width=1\linewidth]{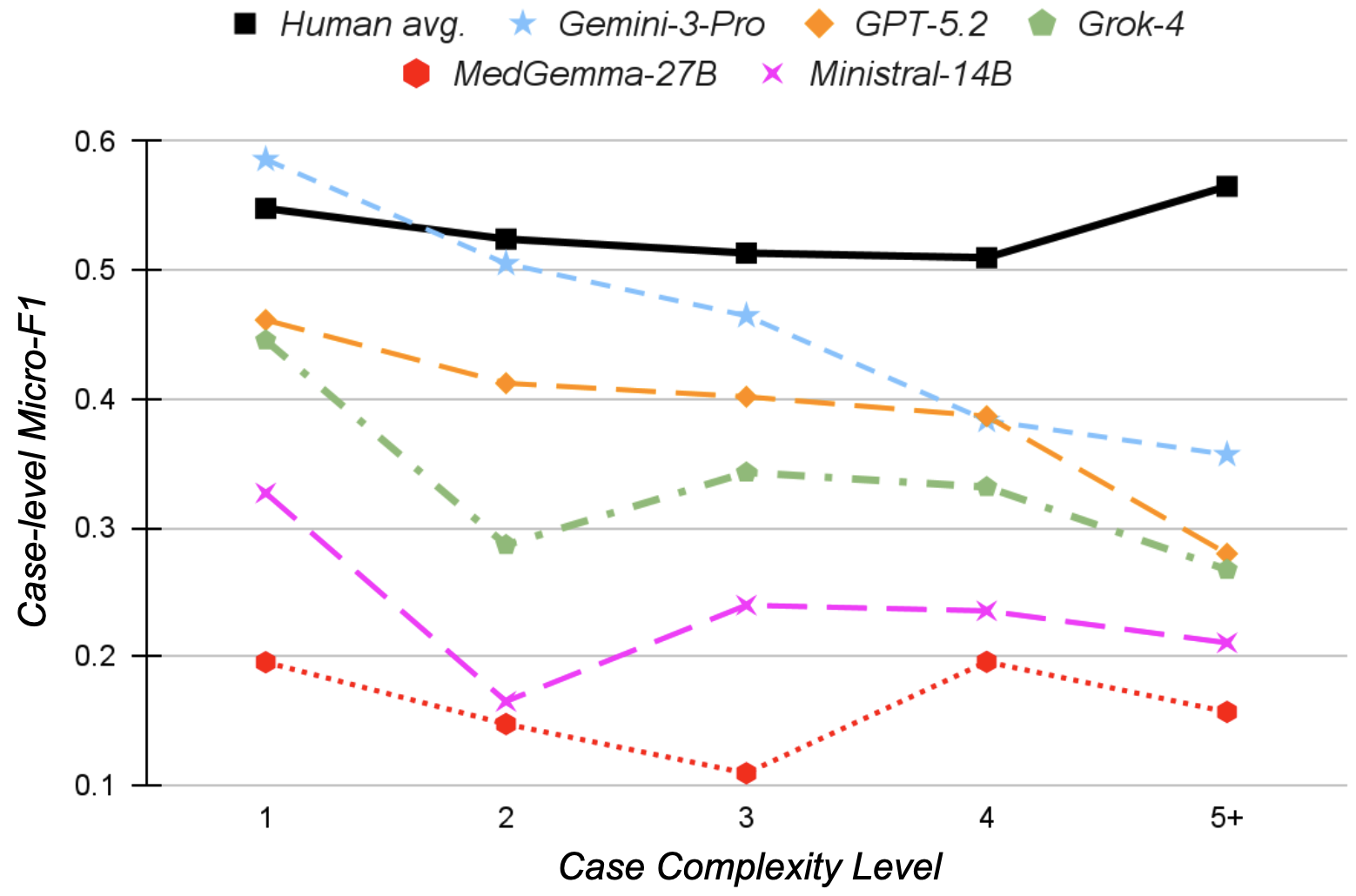}
  \caption{Comparison of case-level Micro-F1 between models and human baseline among different case complexity levels.}
  \label{fig:complexity_vs_performance}
\end{figure}

We then use Gemini-3-Pro as an LLM judge to scale this taxonomy-based analysis to all model outputs. Given the clinician-defined failure descriptions, each model rationale, and the expert-annotated ground-truth rationale, the judge assigns binary labels for the five failure dimensions to each model--case pair; the detailed prompt is provided in Appendix~\ref{subsec:appendix_llm_as_j_prompt}. We apply this procedure to all cases for six representative models and summarize the resulting distributions in Figure~\ref{fig:failure_dimension}. To assess reliability, dental experts manually annotated a random 10\% subset of cases, with three sampled model outputs per case. Agreement between the LLM judge and human experts reached 95.3\%, suggesting that this setup is reliable for large-scale failure analysis. Per-dimension confusion matrices are reported in Appendix~\ref{subsec:appendix_human_agreenment}.

Figure~\ref{fig:failure_dimension} shows that model failures are highly non-uniform across dimensions. The most prevalent failure mode across all six models is \textbf{OPG Finding Omission}, which consistently occurs far more often than any other category, suggesting that current MLLMs more often fail to incorporate clinically relevant radiographic evidence than actively misread it. OPG Misinterpretation is also common, but substantially less frequent, indicating that omission rather than misperception is the dominant radiographic bottleneck on this task.

By contrast, Protocol Misalignment and Hallucination vary more substantially across models and thus account for much of the model-specific failure profile. For example, Claude-Sonnet-4-5 and GPT5.2 exhibit higher rates of protocol misalignment, whereas hallucination is more pronounced in Claude-Sonnet-4-5, Grok-4, and MedGemma-27B. In comparison, Chief Complaint Neglect is rare across all models, suggesting that the main challenge is not simply ignoring the complaint text, but failing to integrate radiographic findings with referral criteria into an appropriate triage decision. Representative case studies are shown in Figure~\ref{fig:failure_patterns_case_study_medgemma} and Appendix~\ref{subsec:appendix_failure_case_study}.

\begin{figure}[t]
  \centering
  \includegraphics[width=1\linewidth]{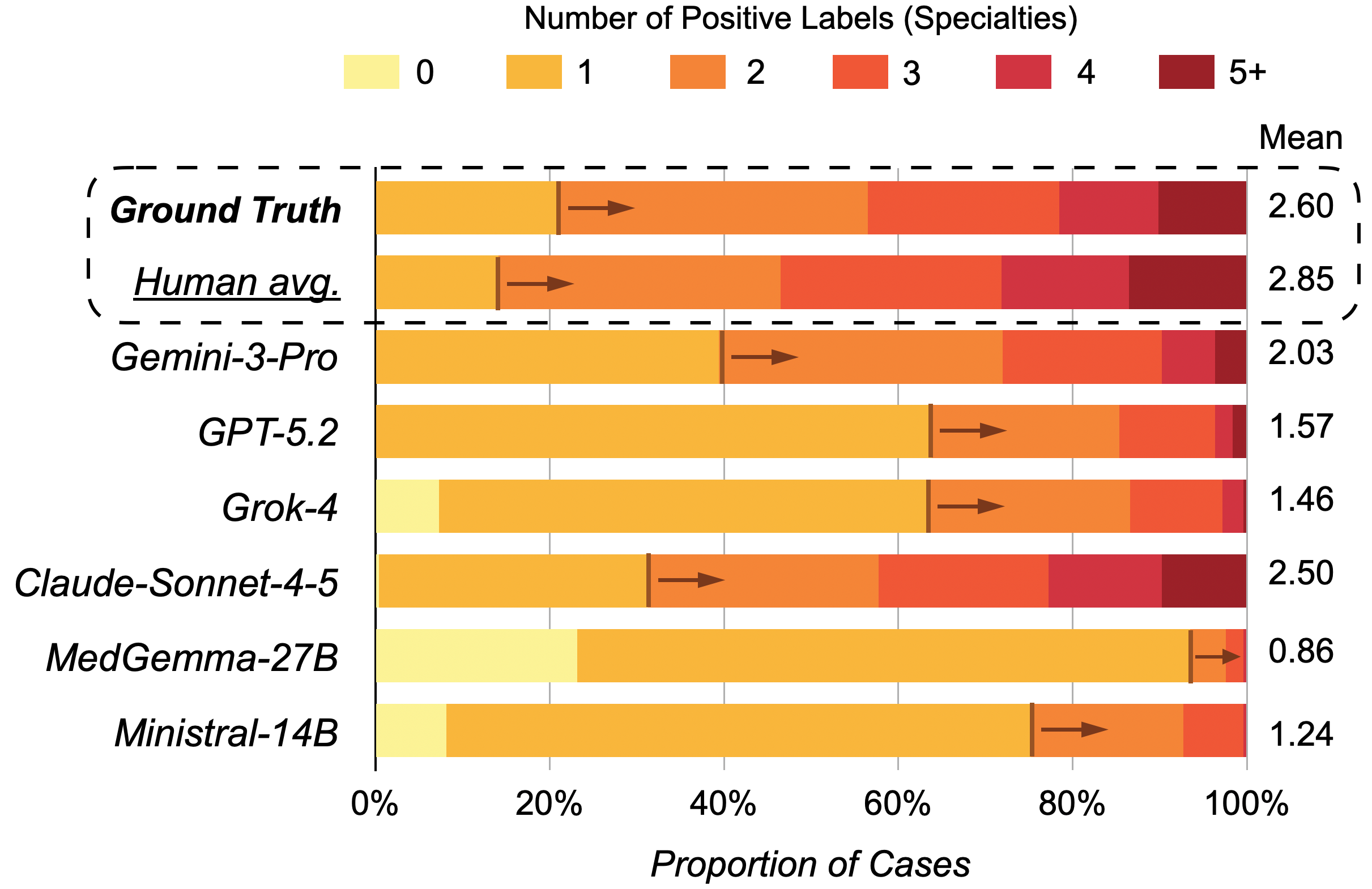}
  \caption{Distribution of the number of positive triage domains per case. The arrow marks the threshold for cases with at least two positive domains.}
  \label{fig:positive_label_dist}
\end{figure}

\subsection{MLLMs Struggle with Complex Triage Cases}
The failure taxonomy above suggests that MLLM errors are not random, but reflect a broader difficulty in turning clinical evidence into appropriate triage decisions. We next show that this limitation is most evident on cases requiring referral coverage across multiple triage domains, where \textit{triage domains} denote the \textit{8 coarse-grained specialty-level referral categories} in our taxonomy and \textit{case complexity} is defined as the number of such domains required by a case in the ground truth, i.e., how many specialty-level referral domains must be covered for that patient.

As case complexity increases, the performance gap between MLLMs and human dentists widens substantially. Figure~\ref{fig:complexity_vs_performance} shows that human performance remains relatively stable, decreasing only from 0.548 on single-domain cases to 0.510 on cases with four positive domains, and recovering to 0.565 on the most complex cases with 5+ positive domains. In contrast, all evaluated MLLMs degrade much more sharply. For example, Gemini-3-Pro drops from 0.586 to 0.357 from single-domain to 5+ domain cases, while GPT-5.2 declines from 0.461 to 0.280. These results suggest that the main challenge in \textsc{Dental-TriageBench} is not merely predicting individual labels, but maintaining adequate referral coverage for multi-problem cases. More broadly, current MLLMs remain substantially weaker than human dentists when triage requires simultaneous reasoning over multiple co-occurring referral needs.

\subsection{MLLMs systematically under-cover referral needs}
To examine whether current MLLMs systematically under-cover referral needs, we next analyze the number of positive coarse-grained triage domains predicted per case. If MLLMs struggle to maintain adequate referral coverage, this should manifest as systematically narrower referral sets than those assigned in ground truth and by human dentists. Figure~\ref{fig:positive_label_dist} shows exactly this pattern: ground truth and human annotations frequently include multiple positive triage domains for the same case, whereas most MLLMs predict substantially fewer. Nearly 80\% of ground-truth and human-annotated cases involve at least two positive triage domains, and about half involve more than two, but such broad referral sets are much less common in model predictions. These results suggest that, as case complexity increases, current MLLMs tend to compress multi-domain referral needs into narrower predictions rather than preserve the breadth of coverage required for realistic triage.

\begin{figure}[t]
  \centering
  \includegraphics[width=1\linewidth]{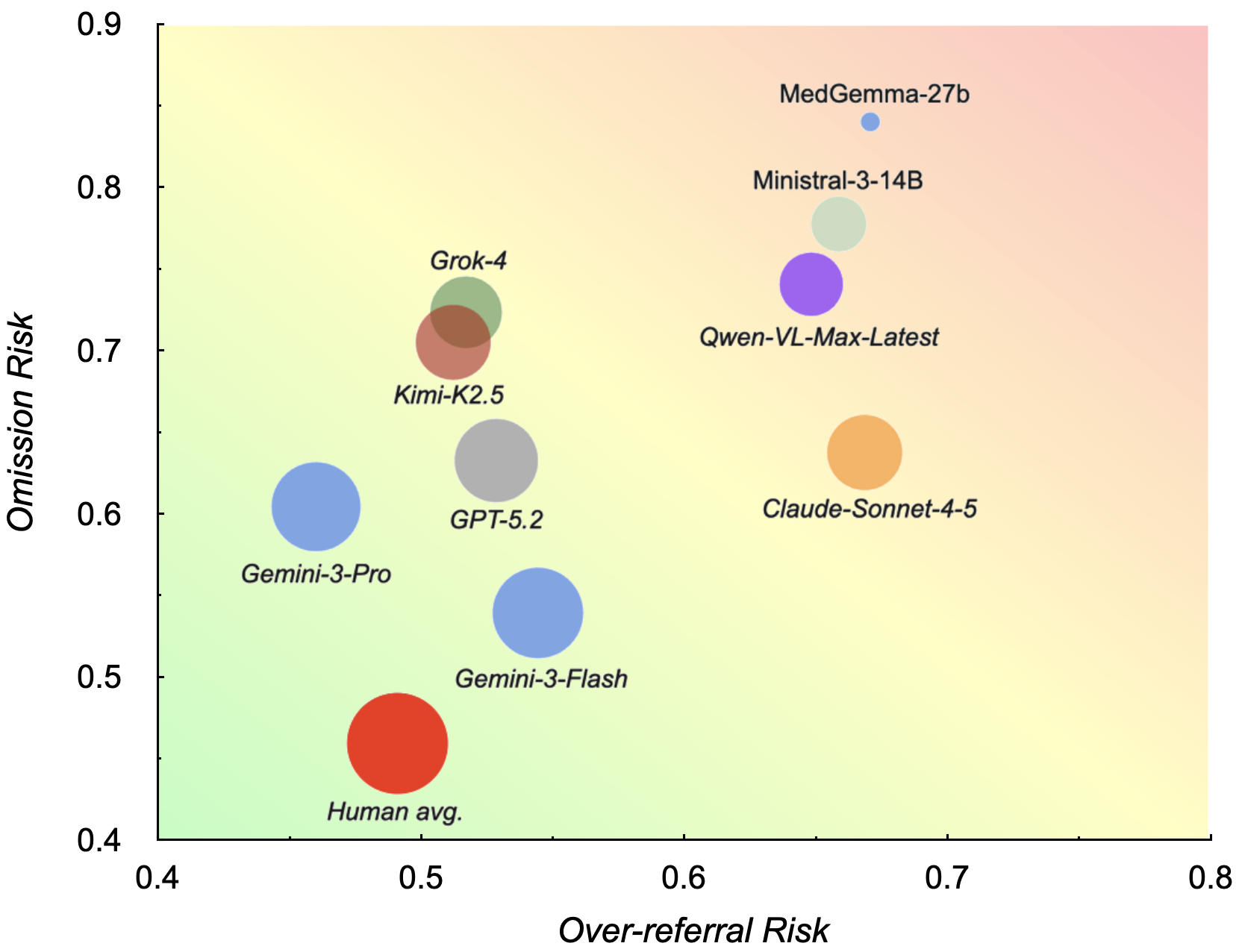}
  \caption{Trade-off between omission risk and over-referral burden. The lower-left region is preferred, indicating fewer missed necessary referrals and fewer unnecessary referrals. Bubble size denotes fine-grained Macro-F1 reported in Table~\ref{tab:merged_fine_coarse_tab}.}
  \label{fig:triage_risk_scatter}
\end{figure}

A key clinical question, however, is whether these narrower referral sets reflect fewer unnecessary referrals or more missed necessary ones. In dental triage, the latter is generally more concerning: while slight over-referral may increase downstream review burden, under-triage can delay appropriate specialty evaluation and leave clinically important needs unaddressed. Figure~\ref{fig:triage_risk_scatter} shows that human raters occupy a region of slightly higher over-referral burden but markedly lower omission risk, whereas MLLMs incur substantially higher omission without a commensurate reduction in over-referral. This suggests that human dentists adopt a safer strategy by preserving broader referral coverage under uncertainty, whereas current MLLMs more often sacrifice necessary referrals without achieving a clear efficiency benefit. In this sense, the narrower referral sets produced by MLLMs reflect not a desirable conservative strategy, but a clinically riskier form of under-coverage.

\section{Conclusion}
We introduce \textsc{Dental-TriageBench}, the first expert-annotated benchmark for reasoning-driven multimodal dental triage from chief complaints and OPGs, and show across 19 MLLMs that current systems remain substantially below human performance, especially on fine-grained referral planning, with clinically concerning omission-heavy under-coverage on complex multi-domain cases. These findings point to several directions for future work: developing coverage-aware triage models that preserve referral completeness under multi-problem uncertainty; improving OPG grounding through dental-specific multimodal pretraining, structured radiographic supervision, or finding-first pipelines; aligning model decisions more tightly with real triage protocols; and advancing uncertainty-aware clinical AI through abstention, selective deferral, and safety-oriented evaluation. We hope \textsc{Dental-TriageBench} will support the development of multimodal dental AI systems that are more clinically grounded, protocol-aligned, coverage-aware, and safe for downstream care.

\clearpage
\section{Limitations}
\textsc{Dental-TriageBench} has several limitations. It is relatively small in scale and collected from a single clinical context, which may limit generalizability. The benchmark is based on OPG--complaint pairs and does not capture the full evidence available in chairside diagnosis. Lastly, our rationale-based failure analysis, although expert-informed and human-verified, is still an approximation of model behavior.

\section{Ethical Considerations}
Dental triage is a safety-critical task, and this work should not be interpreted as evidence for autonomous clinical deployment. Instead, \textsc{Dental-TriageBench} is intended to expose the current limitations of MLLMs under realistic multimodal triage conditions. As our results show, even the strongest models remain below human performance and exhibit omission-heavy errors that may be clinically unsafe, particularly in complex cross-domain cases. All benchmark cases were retrospectively collected from authentic clinical workflows and manually de-identified prior to annotation, with personal identifiers removed from both text and OPG images; all labels were reviewed by licensed senior dentists. Because the benchmark is derived from real clinical data, its use should remain subject to appropriate privacy, institutional, and data-governance requirements. More broadly, benchmark performance should not be conflated with clinical competence, since real-world triage depends on additional examination findings and contextual information beyond chief complaints and OPGs. We therefore position \textsc{Dental-TriageBench} as a research benchmark for developing safer and more clinically grounded decision-support systems, rather than as a substitute for qualified dental professionals.


\nocite{*}
\bibliography{ref}

@article{van_nistelrooij_combining_2024,
	title = {Combining public datasets for automated tooth assessment in panoramic radiographs},
	volume = {24},
	issn = {1472-6831},
	url = {https://pubmed.ncbi.nlm.nih.gov/38532414/},
	doi = {10.1186/s12903-024-04129-5},
	number = {1},
	journal = {BMC Oral Health},
	author = {van Nistelrooij, N. and Ghoul, K. E. and Xi, T. and Saha, A. and Kempers, S. and Cenci, M. and Loomans, B. and Flügge, T. and van Ginneken, B. and Vinayahalingam, S.},
	month = mar,
	year = {2024},
	pmid = {38532414},
	pmcid = {PMC10964594},
	pages = {387},
}

@misc{jang2026omnidentaccessibleexplainableai,
      title={OMNI-Dent: Towards an Accessible and Explainable AI Framework for Automated Dental Diagnosis}, 
      author={Leeje Jang and Yao-Yi Chiang and Angela M. Hastings and Patimaporn Pungchanchaikul and Martha B. Lucas and Emily C. Schultz and Jeffrey P. Louie and Mohamed Estai and Wen-Chen Wang and Ryan H. L. Ip and Boyen Huang},
      year={2026},
      eprint={2602.07041},
      archivePrefix={arXiv},
      primaryClass={cs.CV},
      url={https://arxiv.org/abs/2602.07041}, 
}

@misc{yu2026opgagentagentauditabledental,
      title={OPGAgent: An Agent for Auditable Dental Panoramic X-ray Interpretation}, 
      author={Zhaolin Yu and Litao Yang and Ben Babicka and Ming Hu and Jing Hao and Anthony Huang and James Huang and Yueming Jin and Jiasong Wu and Zongyuan Ge},
      year={2026},
      eprint={2603.00462},
      archivePrefix={arXiv},
      primaryClass={cs.CV},
      url={https://arxiv.org/abs/2603.00462}, 
}

@article{Li2024,
    author    = {Li, X. and Ma, X. and Zhao, Y. and Hu, J. and Liu, J. and Yang, Z. and Han, F. and Zhang, J. and Liu, W. and Zhou, Z.},
    title     = {A Multi-center Dental Panoramic Radiography Image Dataset for Impacted Teeth, Periodontitis, and Dental Caries: Benchmarking Segmentation and Classification Tasks},
    journal   = {Journal of Imaging Informatics in Medicine},
    year      = {2024},
    volume    = {37},
    number    = {2},
    pages     = {831--841},
    month     = {apr},
    doi       = {10.1007/s10278-024-00972-8},
    note      = {Epub 2024 Feb 6},
    pmid      = {38321312},
    pmcid     = {PMC11031544}
}

@misc{meng2025dentvlmmultimodalvisionlanguagemodel,
      title={DentVLM: A Multimodal Vision-Language Model for Comprehensive Dental Diagnosis and Enhanced Clinical Practice}, 
      author={Zijie Meng and Jin Hao and Xiwei Dai and Yang Feng and Jiaxiang Liu and Bin Feng and Huikai Wu and Xiaotang Gai and Hengchuan Zhu and Tianxiang Hu and Yangyang Wu and Hongxia Xu and Jin Li and Jun Xiao and Xiaoqiang Liu and Joey Tianyi Zhou and Fudong Zhu and Zhihe Zhao and Lunguo Xia and Bing Fang and Jimeng Sun and Jian Wu and Zuozhu Liu},
      year={2025},
      eprint={2509.23344},
      archivePrefix={arXiv},
      primaryClass={cs.CV},
      url={https://arxiv.org/abs/2509.23344}, 
}

@misc{hao2025betterdentalaimultimodal,
      title={Towards Better Dental AI: A Multimodal Benchmark and Instruction Dataset for Panoramic X-ray Analysis}, 
      author={Jing Hao and Yuxuan Fan and Yanpeng Sun and Kaixin Guo and Lizhuo Lin and Jinrong Yang and Qi Yong H. Ai and Lun M. Wong and Hao Tang and Kuo Feng Hung},
      year={2025},
      eprint={2509.09254},
      archivePrefix={arXiv},
      primaryClass={cs.CV},
      url={https://arxiv.org/abs/2509.09254}, 
}

@misc{huang2025generalistintelligencedentistryvision,
      title={Towards Generalist Intelligence in Dentistry: Vision Foundation Models for Oral and Maxillofacial Radiology}, 
      author={Xinrui Huang and Fan Xiao and Dongming He and Anqi Gao and Dandan Li and Xiaofan Zhang and Shaoting Zhang and Xudong Wang},
      year={2025},
      eprint={2510.14532},
      archivePrefix={arXiv},
      primaryClass={cs.CV},
      url={https://arxiv.org/abs/2510.14532}, 
}

@inproceedings{liu2024clinicbench,
  author       = {Fenglin Liu and
                  Zheng Li and
                  Hongjian Zhou and
                  Qingyu Yin and
                  Jingfeng Yang and
                  Xianfeng Tang and
                  Chen Luo and
                  Ming Zeng and
                  Haoming Jiang and
                  Yifan Gao and
                  Priyanka Nigam and
                  Sreyashi Nag and
                  Bing Yin and
                  Yining Hua and
                  Xuan Zhou and
                  Omid Rohanian and
                  Anshul Thakur and
                  Lei A. Clifton and
                  David A. Clifton},
  editor       = {Yaser Al{-}Onaizan and
                  Mohit Bansal and
                  Yun{-}Nung Chen},
  title        = {Large Language Models Are Poor Clinical Decision-Makers: {A} Comprehensive
                  Benchmark},
  booktitle    = {Proceedings of the 2024 Conference on Empirical Methods in Natural
                  Language Processing, {EMNLP} 2024, Miami, FL, USA, November 12-16,
                  2024},
  pages        = {13696--13710},
  publisher    = {Association for Computational Linguistics},
  year         = {2024},
  url          = {https://doi.org/10.18653/v1/2024.emnlp-main.759},
  doi          = {10.18653/V1/2024.EMNLP-MAIN.759},
  bibsource    = {dblp computer science bibliography, https://dblp.org}
}

@inproceedings{ouyang2024climedbench,
  author       = {Zetian Ouyang and
                  Yishuai Qiu and
                  Linlin Wang and
                  Gerard de Melo and
                  Ya Zhang and
                  Yanfeng Wang and
                  Liang He},
  editor       = {Yaser Al{-}Onaizan and
                  Mohit Bansal and
                  Yun{-}Nung Chen},
  title        = {CliMedBench: {A} Large-Scale Chinese Benchmark for Evaluating Medical
                  Large Language Models in Clinical Scenarios},
  booktitle    = {Proceedings of the 2024 Conference on Empirical Methods in Natural
                  Language Processing, {EMNLP} 2024, Miami, FL, USA, November 12-16,
                  2024},
  pages        = {8428--8438},
  publisher    = {Association for Computational Linguistics},
  year         = {2024},
  url          = {https://doi.org/10.18653/v1/2024.emnlp-main.480},
  doi          = {10.18653/V1/2024.EMNLP-MAIN.480},
  bibsource    = {dblp computer science bibliography, https://dblp.org}
}

@inproceedings{zhang2025llmevalmed,
  author       = {Ming Zhang and
                  Yujiong Shen and
                  Zelin Li and
                  Huayu Sha and
                  Binze Hu and
                  Yuhui Wang and
                  Chenhao Huang and
                  Shichun Liu and
                  Jingqi Tong and
                  Changhao Jiang and
                  Mingxu Chai and
                  Zhiheng Xi and
                  Shihan Dou and
                  Tao Gui and
                  Qi Zhang and
                  Xuanjing Huang},
  editor       = {Christos Christodoulopoulos and
                  Tanmoy Chakraborty and
                  Carolyn Rose and
                  Violet Peng},
  title        = {LLMEval-Med: {A} Real-world Clinical Benchmark for Medical LLMs with
                  Physician Validation},
  booktitle    = {Findings of the Association for Computational Linguistics: {EMNLP}
                  2025, Suzhou, China, November 4-9, 2025},
  pages        = {4888--4914},
  publisher    = {Association for Computational Linguistics},
  year         = {2025},
  url          = {https://aclanthology.org/2025.findings-emnlp.263/},
  bibsource    = {dblp computer science bibliography, https://dblp.org}
}

@inproceedings{lu2024triageagent,
  author       = {Meng Lu and
                  Brandon Ho and
                  Dennis Ren and
                  Xuan Wang},
  editor       = {Yaser Al{-}Onaizan and
                  Mohit Bansal and
                  Yun{-}Nung Chen},
  title        = {TriageAgent: Towards Better Multi-Agents Collaborations for Large
                  Language Model-Based Clinical Triage},
  booktitle    = {Findings of the Association for Computational Linguistics: {EMNLP}
                  2024, Miami, Florida, USA, November 12-16, 2024},
  series       = {Findings of {ACL}},
  volume       = {{EMNLP} 2024},
  pages        = {5747--5764},
  publisher    = {Association for Computational Linguistics},
  year         = {2024},
  url          = {https://doi.org/10.18653/v1/2024.findings-emnlp.329},
  doi          = {10.18653/V1/2024.FINDINGS-EMNLP.329},
  bibsource    = {dblp computer science bibliography, https://dblp.org}
}

@article{williams2024clinicalacuity,
  title={Use of a large language model to assess clinical acuity of adults in the emergency department},
  author={Williams, Christopher YK and Zack, Travis and Miao, Brenda Y and Sushil, Madhumita and Wang, Michelle and Kornblith, Aaron E and Butte, Atul J},
  journal={JAMA network open},
  volume={7},
  number={5},
  pages={e248895},
  year={2024}
}

@article{masanneck2024triageperformance,
  title={Triage performance across large language models, ChatGPT, and untrained doctors in emergency medicine: comparative study},
  author={Masanneck, Lars and Schmidt, Linea and Seifert, Antonia and K{\"o}lsche, Tristan and Huntemann, Niklas and Jansen, Robin and Mehsin, Mohammed and Bernhard, Michael and Meuth, Sven G and B{\"o}hm, Lennert and others},
  journal={Journal of medical Internet research},
  volume={26},
  pages={e53297},
  year={2024},
  publisher={JMIR Publications Toronto, Canada}
}

@inproceedings{li2023llavamed,
  author       = {Chunyuan Li and
                  Cliff Wong and
                  Sheng Zhang and
                  Naoto Usuyama and
                  Haotian Liu and
                  Jianwei Yang and
                  Tristan Naumann and
                  Hoifung Poon and
                  Jianfeng Gao},
  editor       = {Alice Oh and
                  Tristan Naumann and
                  Amir Globerson and
                  Kate Saenko and
                  Moritz Hardt and
                  Sergey Levine},
  title        = {LLaVA-Med: Training a Large Language-and-Vision Assistant for Biomedicine
                  in One Day},
  booktitle    = {Advances in Neural Information Processing Systems 36: Annual Conference
                  on Neural Information Processing Systems 2023, NeurIPS 2023, New Orleans,
                  LA, USA, December 10 - 16, 2023},
  year         = {2023},
  url          = {http://papers.nips.cc/paper\_files/paper/2023/hash/5abcdf8ecdcacba028c6662789194572-Abstract-Datasets\_and\_Benchmarks.html},
  bibsource    = {dblp computer science bibliography, https://dblp.org}
}

@inproceedings{moor2023medflamingo,
  author       = {Michael Moor and
                  Qian Huang and
                  Shirley Wu and
                  Michihiro Yasunaga and
                  Yash Dalmia and
                  Jure Leskovec and
                  Cyril Zakka and
                  Eduardo Pontes Reis and
                  Pranav Rajpurkar},
  editor       = {Stefan Hegselmann and
                  Antonio Parziale and
                  Divya Shanmugam and
                  Shengpu Tang and
                  Mercy Nyamewaa Asiedu and
                  Serina Chang and
                  Tom Hartvigsen and
                  Harvineet Singh},
  title        = {Med-Flamingo: a Multimodal Medical Few-shot Learner},
  booktitle    = {Machine Learning for Health, ML4H@NeurIPS 2023, 10 December 2023,
                  New Orleans, Louisiana, {USA}},
  series       = {Proceedings of Machine Learning Research},
  volume       = {225},
  pages        = {353--367},
  publisher    = {{PMLR}},
  year         = {2023},
  url          = {https://proceedings.mlr.press/v225/moor23a.html},
  bibsource    = {dblp computer science bibliography, https://dblp.org}
}

@article{zhang2024biomedgpt,
  title={A generalist vision--language foundation model for diverse biomedical tasks},
  author={Zhang, Kai and Zhou, Rong and Adhikarla, Eashan and Yan, Zhiling and Liu, Yixin and Yu, Jun and Liu, Zhengliang and Chen, Xun and Davison, Brian D and Ren, Hui and others},
  journal={Nature medicine},
  volume={30},
  number={11},
  pages={3129--3141},
  year={2024},
  publisher={Nature Publishing Group US New York}
}

@article{tu2023generalistbiomedical,
  title={Towards generalist biomedical AI},
  author={Tu, Tao and Azizi, Shekoofeh and Driess, Danny and Schaekermann, Mike and Amin, Mohamed and Chang, Pi-Chuan and Carroll, Andrew and Lau, Charles and Tanno, Ryutaro and Ktena, Ira and others},
  journal={Nejm Ai},
  volume={1},
  number={3},
  pages={AIoa2300138},
  year={2024},
  publisher={Massachusetts Medical Society}
}

@article{chen2024chexagent,
  title={A vision-language foundation model to enhance efficiency of chest X-ray interpretation},
  author={Chen, Zhihong and Varma, Maya and Xu, Justin and Paschali, Magdalini and Van Veen, Dave and Johnston, Andrew and Youssef, Alaa and Blankemeier, Louis and Bluethgen, Christian and Altmayer, Stephan and others},
  journal={arXiv preprint arXiv:2401.12208},
  year={2024}
}

@inproceedings{gmai2024mmbench,
  author       = {Pengcheng Chen and
                  Jin Ye and
                  Guoan Wang and
                  Yanjun Li and
                  Zhongying Deng and
                  Wei Li and
                  Tianbin Li and
                  Haodong Duan and
                  Ziyan Huang and
                  Yanzhou Su and
                  Benyou Wang and
                  Shaoting Zhang and
                  Bin Fu and
                  Jianfei Cai and
                  Bohan Zhuang and
                  Eric J. Seibel and
                  Junjun He and
                  Yu Qiao},
  editor       = {Amir Globersons and
                  Lester Mackey and
                  Danielle Belgrave and
                  Angela Fan and
                  Ulrich Paquet and
                  Jakub M. Tomczak and
                  Cheng Zhang},
  title        = {GMAI-MMBench: {A} Comprehensive Multimodal Evaluation Benchmark Towards
                  General Medical {AI}},
  booktitle    = {Advances in Neural Information Processing Systems 38: Annual Conference
                  on Neural Information Processing Systems 2024, NeurIPS 2024, Vancouver,
                  BC, Canada, December 10 - 15, 2024},
  year         = {2024},
  url          = {http://papers.nips.cc/paper\_files/paper/2024/hash/ab7e02fd60e47e2a379d567f6b54f04e-Abstract-Datasets\_and\_Benchmarks\_Track.html},
  bibsource    = {dblp computer science bibliography, https://dblp.org}
}

@inproceedings{
zhou2025drvdbench,
title={Dr{VD}-Bench: Do Vision-Language Models Reason Like Human Doctors in Medical Image Diagnosis?},
author={Tianhong Zhou and Yin Xu and Yingtao Zhu and Chuxi Xiao and Haiyang Bian and Lei Wei and Xuegong Zhang},
booktitle={The Thirty-ninth Annual Conference on Neural Information Processing Systems Datasets and Benchmarks Track},
year={2025},
url={https://openreview.net/forum?id=fqpR9FRso9}
}

@article{estai2022toothnumbering,
  title={Deep learning for automated detection and numbering of permanent teeth on panoramic images},
  author={Estai, Mohamed and Tennant, Marc and Gebauer, Dieter and Brostek, Andrew and Vignarajan, Janardhan and Mehdizadeh, Maryam and Saha, Sajib},
  journal={Dentomaxillofacial Radiology},
  volume={51},
  number={2},
  pages={20210296},
  year={2022},
  publisher={Oxford University Press}
}

@article{zhu2023multipledentaldiseases,
  title={Artificial intelligence in the diagnosis of dental diseases on panoramic radiographs: a preliminary study},
  author={Zhu, Junhua and Chen, Zhi and Zhao, Jing and Yu, Yueyuan and Li, Xiaojuan and Shi, Kangjian and Zhang, Fan and Yu, Feifei and Shi, Keying and Sun, Zhe and others},
  journal={BMC Oral Health},
  volume={23},
  number={1},
  pages={358},
  year={2023},
  publisher={Springer}
}

@article{turosz2023overview,
  title={Applications of artificial intelligence in the analysis of dental panoramic radiographs: an overview of systematic reviews},
  author={Turosz, Natalia and Ch{\k{e}}ci{\'n}ska, Kamila and Ch{\k{e}}ci{\'n}ski, Maciej and Brzozowska, Anita and Nowak, Zuzanna and Sikora, Maciej},
  journal={Dentomaxillofacial Radiology},
  volume={52},
  number={7},
  pages={20230284},
  year={2023},
  publisher={Oxford University Press}
}

@article{wang2025multinationaldpr,
  title={Artificial Intelligence to Assess Dental Findings from Panoramic Radiographs--A Multinational Study},
  author={Wang, Yin-Chih Chelsea and Chen, Tsao-Lun and Vinayahalingam, Shankeeth and Wu, Tai-Hsien and Chang, Chu Wei and Chang, Hsuan Hao and Wei, Hung-Jen and Chen, Mu-Hsiung and Ko, Ching-Chang and Moin, David Anssari and others},
  journal={arXiv preprint arXiv:2502.10277},
  year={2025}
}

@inproceedings{
hao2025mmoral,
title={Towards Better Dental {AI}: A Multimodal Benchmark and Instruction Dataset for Panoramic X-ray Analysis},
author={Jing Hao and Yuxuan Fan and Yanpeng Sun and Kaixin Guo and Lin Lizhuo and Jinrong Yang and Qiyong Hemis Ai and Lun M Wong and Hao Tang and Kuo Feng Hung},
booktitle={The Thirty-ninth Annual Conference on Neural Information Processing Systems Datasets and Benchmarks Track},
year={2025},
url={https://openreview.net/forum?id=nxGSj1xkm3}
}

@article{meng2025dentvlm,
  title={Dentvlm: A multimodal vision-language model for comprehensive dental diagnosis and enhanced clinical practice},
  author={Meng, Zijie and Hao, Jin and Dai, Xiwei and Feng, Yang and Liu, Jiaxiang and Feng, Bin and Wu, Huikai and Gai, Xiaotang and Zhu, Hengchuan and Hu, Tianxiang and others},
  journal={arXiv preprint arXiv:2509.23344},
  year={2025}
}

@article{huang2025dentvfm,
  title={Towards Generalist Intelligence in Dentistry: Vision Foundation Models for Oral and Maxillofacial Radiology},
  author={Huang, Xinrui and Xiao, Fan and He, Dongming and Gao, Anqi and Li, Dandan and Zhang, Xiaofan and Zhang, Shaoting and Wang, Xudong},
  journal={arXiv preprint arXiv:2510.14532},
  year={2025}
}

@inproceedings{jang2026omnident,
  title={OMNI-Dent: Towards an Accessible and Explainable AI Framework for Automated Dental Diagnosis},
  author={Jang, Leeje and Chiang, Yao-Yi and Hastings, Angela M and Pungchanchaikul, Patimaporn and Lucas, Martha B and Schultz, Emily C and Louie, Jeffrey P and Estai, Mohamed and Wang, Wen-Chen and Ip, Ryan HL and others},
  booktitle={Proceedings of the IEEE/CVF Winter Conference on Applications of Computer Vision},
  pages={415--424},
  year={2026}
}

@article{balel2026hybridopgreports,
  title={A Novel Hybrid Large Language Model Approach for Reporting Panoramic Radiographs and Performance Comparison with Current Large Language Models},
  author={Balel, Yunus and Sa{\u{g}}ta{\c{s}}, Kaan and Teke, Fatih and Kurt, Mehmet Ali},
  journal={Journal of imaging informatics in medicine},
  year={2026}
}

@article{demograph,
  author       = {Yushi Feng and
                  Tsai Hor Chan and
                  Guosheng Yin and
                  Lequan Yu},
  title        = {Democratizing large language model-based graph data augmentation via
                  latent knowledge graphs},
  journal      = {Neural Networks},
  volume       = {191},
  pages        = {107777},
  year         = {2025},
  url          = {https://doi.org/10.1016/j.neunet.2025.107777},
  doi          = {10.1016/J.NEUNET.2025.107777},
  bibsource    = {dblp computer science bibliography, https://dblp.org}
}

@inproceedings{
chen2025ipad,
title={{IPAD}: Inverse Prompt for {AI} Detection - A Robust and Interpretable {LLM}-Generated Text Detector},
author={Zheng Chen and Yushi Feng and Jisheng Dang and Changyang He and Yue Deng and Hongxi Pu and Haoxuan Li and Bo Li},
booktitle={The Thirty-ninth Annual Conference on Neural Information Processing Systems},
year={2025},
url={https://openreview.net/forum?id=3JoQTGhUzz}
}

@inproceedings{feng2026pass,
  title     = {PASS: Probabilistic Agentic Supernet Sampling for Interpretable and Adaptive Chest X-Ray Reasoning},
  author    = {Feng, Yushi and Du, Junye and Hong, Yingying and Wang, Qifan and Yu, Lequan},
  booktitle = {Proceedings of the AAAI Conference on Artificial Intelligence},
  year      = {2026},
  note      = {AAAI-26}
}

@inproceedings{vqa,
  author       = {Ziyan Xiao and
                  Ruiyang Zhang and
                  Yushi Feng and
                  Lingting Zhu and
                  Liang Peng and
                  Lequan Yu},
  title        = {A Dynamic Agent Framework for Large Language Model Reasoning for Medical
                  and Visual Question Answering},
  booktitle    = {{IEEE/CVF} International Conference on Computer Vision, {ICCV} 2025
                  - Workshops, Honolulu, HI, USA, October 19-20, 2025},
  pages        = {1154--1163},
  publisher    = {{IEEE}},
  year         = {2025},
  url          = {https://doi.org/10.1109/ICCVW69036.2025.00124},
  doi          = {10.1109/ICCVW69036.2025.00124},
  bibsource    = {dblp computer science bibliography, https://dblp.org}
}

@misc{anthropic2025claude45,
  author = {{Anthropic}},
  title = {Introducing Claude Sonnet 4.5},
  year = {2025},
  url = {https://www.anthropic.com/news/claude-sonnet-4-5},
  note = {Official announcement}
}

@misc{google2025gemini2flash,
  author = {{Google}},
  title = {Gemini 2 Flash | Gemini API | Google AI for Developers},
  year = {2025},
  url = {https://ai.google.dev/gemini-api/docs/models/gemini-2.0-flash},
  note = {Official developer documentation}
}

@misc{google2026gemini3flashpreview,
  author = {{Google}},
  title = {Gemini 3 Flash Preview | Gemini API | Google AI for Developers},
  year = {2026},
  url = {https://ai.google.dev/gemini-api/docs/models/gemini-3-flash-preview},
  note = {Official developer documentation}
}

@misc{google2026gemini3propreview,
  author = {{Google}},
  title = {Gemini 3 Pro Preview | Gemini API | Google AI for Developers},
  year = {2026},
  url = {https://ai.google.dev/gemini-api/docs/models/gemini-3-pro-preview},
  note = {Official developer documentation}
}

@misc{openai2025gpt52,
  author = {{OpenAI}},
  title = {Introducing GPT-5.2},
  year = {2025},
  url = {https://openai.com/index/introducing-gpt-5-2/},
  note = {Official announcement}
}

@misc{xai2025grok4,
  author = {{xAI}},
  title = {Grok 4},
  year = {2025},
  url = {https://x.ai/news/grok-4},
  note = {Official announcement}
}

@article{kimi2026k25,
  title = {Kimi K2.5: Visual Agentic Intelligence},
  author = {{Kimi}},
  journal = {arXiv preprint arXiv:2602.02276},
  year = {2026}
}

@misc{alibaba2026qwenvlmax,
  author = {{Alibaba}},
  title = {OpenAI-compatible Vision API for Qwen-VL Models},
  year = {2026},
  url = {https://help.aliyun.com/zh/model-studio/qwen-vl-compatible-with-openai},
  note = {Official developer documentation for qwen-vl-max / qwen-vl-max-latest}
}

@article{glmv2025glm41vthinking,
  title = {GLM-4.1V-Thinking and GLM-4.5V: Towards Versatile Multimodal Reasoning with Scalable Reinforcement Learning},
  author = {{GLM-V Team}},
  journal = {arXiv preprint arXiv:2507.01006},
  year = {2025}
}

@misc{wang2025internvl35,
      title={InternVL3.5: Advancing Open-Source Multimodal Models in Versatility, Reasoning, and Efficiency}, 
      author={Weiyun Wang and Zhangwei Gao and Lixin Gu and Hengjun Pu and Long Cui and Xingguang Wei and Zhaoyang Liu and Linglin Jing and Shenglong Ye and Jie Shao and Zhaokai Wang and Zhe Chen and Hongjie Zhang and Ganlin Yang and Haomin Wang and Qi Wei and Jinhui Yin and Wenhao Li and Erfei Cui and Guanzhou Chen and Zichen Ding and Changyao Tian and Zhenyu Wu and Jingjing Xie and Zehao Li and Bowen Yang and Yuchen Duan and Xuehui Wang and Zhi Hou and Haoran Hao and Tianyi Zhang and Songze Li and Xiangyu Zhao and Haodong Duan and Nianchen Deng and Bin Fu and Yinan He and Yi Wang and Conghui He and Botian Shi and Junjun He and Yingtong Xiong and Han Lv and Lijun Wu and Wenqi Shao and Kaipeng Zhang and Huipeng Deng and Biqing Qi and Jiaye Ge and Qipeng Guo and Wenwei Zhang and Songyang Zhang and Maosong Cao and Junyao Lin and Kexian Tang and Jianfei Gao and Haian Huang and Yuzhe Gu and Chengqi Lyu and Huanze Tang and Rui Wang and Haijun Lv and Wanli Ouyang and Limin Wang and Min Dou and Xizhou Zhu and Tong Lu and Dahua Lin and Jifeng Dai and Weijie Su and Bowen Zhou and Kai Chen and Yu Qiao and Wenhai Wang and Gen Luo},
      year={2025},
      eprint={2508.18265},
      archivePrefix={arXiv},
      primaryClass={cs.CV},
      url={https://arxiv.org/abs/2508.18265}, 
}

@article{xiaomi2025mimovl,
  title = {MiMo-VL Technical Report},
  author = {{Xiaomi}},
  journal = {arXiv preprint arXiv:2506.03569},
  year = {2025}
}

@misc{liu2026ministral3,
  title={Ministral 3}, 
  author={Alexander H. Liu and Kartik Khandelwal and Sandeep Subramanian and Victor Jouault and Abhinav Rastogi and Adrien Sadé and Alan Jeffares and Albert Jiang and Alexandre Cahill and Alexandre Gavaudan and Alexandre Sablayrolles and Amélie Héliou and Amos You and Andy Ehrenberg and Andy Lo and Anton Eliseev and Antonia Calvi and Avinash Sooriyarachchi and Baptiste Bout and Baptiste Rozière and Baudouin De Monicault and Clémence Lanfranchi and Corentin Barreau and Cyprien Courtot and Daniele Grattarola and Darius Dabert and Diego de las Casas and Elliot Chane-Sane and Faruk Ahmed and Gabrielle Berrada and Gaëtan Ecrepont and Gauthier Guinet and Georgii Novikov and Guillaume Kunsch and Guillaume Lample and Guillaume Martin and Gunshi Gupta and Jan Ludziejewski and Jason Rute and Joachim Studnia and Jonas Amar and Joséphine Delas and Josselin Somerville Roberts and Karmesh Yadav and Khyathi Chandu and Kush Jain and Laurence Aitchison and Laurent Fainsin and Léonard Blier and Lingxiao Zhao and Louis Martin and Lucile Saulnier and Luyu Gao and Maarten Buyl and Margaret Jennings and Marie Pellat and Mark Prins and Mathieu Poirée and Mathilde Guillaumin and Matthieu Dinot and Matthieu Futeral and Maxime Darrin and Maximilian Augustin and Mia Chiquier and Michel Schimpf and Nathan Grinsztajn and Neha Gupta and Nikhil Raghuraman and Olivier Bousquet and Olivier Duchenne and Patricia Wang and Patrick von Platen and Paul Jacob and Paul Wambergue and Paula Kurylowicz and Pavankumar Reddy Muddireddy and Philomène Chagniot and Pierre Stock and Pravesh Agrawal and Quentin Torroba and Romain Sauvestre and Roman Soletskyi and Rupert Menneer and Sagar Vaze and Samuel Barry and Sanchit Gandhi and Siddhant Waghjale and Siddharth Gandhi and Soham Ghosh and Srijan Mishra and Sumukh Aithal and Szymon Antoniak and Teven Le Scao and Théo Cachet and Theo Simon Sorg and Thibaut Lavril and Thiziri Nait Saada and Thomas Chabal and Thomas Foubert and Thomas Robert and Thomas Wang and Tim Lawson and Tom Bewley and Tom Bewley and Tom Edwards and Umar Jamil and Umberto Tomasini and Valeriia Nemychnikova and Van Phung and Vincent Maladière and Virgile Richard and Wassim Bouaziz and Wen-Ding Li and William Marshall and Xinghui Li and Xinyu Yang and Yassine El Ouahidi and Yihan Wang and Yunhao Tang and Zaccharie Ramzi},
  year={2026},
  eprint={2601.08584},
  archivePrefix={arXiv},
  primaryClass={cs.CL},
  url={https://arxiv.org/abs/2601.08584}, 
}

@misc{bai2025qwen25vl,
  title={Qwen2.5-VL Technical Report}, 
  author={Shuai Bai and Keqin Chen and Xuejing Liu and Jialin Wang and Wenbin Ge and Sibo Song and Kai Dang and Peng Wang and Shijie Wang and Jun Tang and Humen Zhong and Yuanzhi Zhu and Mingkun Yang and Zhaohai Li and Jianqiang Wan and Pengfei Wang and Wei Ding and Zheren Fu and Yiheng Xu and Jiabo Ye and Xi Zhang and Tianbao Xie and Zesen Cheng and Hang Zhang and Zhibo Yang and Haiyang Xu and Junyang Lin},
  year={2025},
  eprint={2502.13923},
  archivePrefix={arXiv},
  primaryClass={cs.CV},
  url={https://arxiv.org/abs/2502.13923}, 
}

@misc{bai2025qwen3vl,
  title={Qwen3-VL Technical Report}, 
  author={Shuai Bai and Yuxuan Cai and Ruizhe Chen and Keqin Chen and Xionghui Chen and Zesen Cheng and Lianghao Deng and Wei Ding and Chang Gao and Chunjiang Ge and Wenbin Ge and Zhifang Guo and Qidong Huang and Jie Huang and Fei Huang and Binyuan Hui and Shutong Jiang and Zhaohai Li and Mingsheng Li and Mei Li and Kaixin Li and Zicheng Lin and Junyang Lin and Xuejing Liu and Jiawei Liu and Chenglong Liu and Yang Liu and Dayiheng Liu and Shixuan Liu and Dunjie Lu and Ruilin Luo and Chenxu Lv and Rui Men and Lingchen Meng and Xuancheng Ren and Xingzhang Ren and Sibo Song and Yuchong Sun and Jun Tang and Jianhong Tu and Jianqiang Wan and Peng Wang and Pengfei Wang and Qiuyue Wang and Yuxuan Wang and Tianbao Xie and Yiheng Xu and Haiyang Xu and Jin Xu and Zhibo Yang and Mingkun Yang and Jianxin Yang and An Yang and Bowen Yu and Fei Zhang and Hang Zhang and Xi Zhang and Bo Zheng and Humen Zhong and Jingren Zhou and Fan Zhou and Jing Zhou and Yuanzhi Zhu and Ke Zhu},
  year={2025},
  eprint={2511.21631},
  archivePrefix={arXiv},
  primaryClass={cs.CV},
  url={https://arxiv.org/abs/2511.21631}, 
}

@article{li2024llavaonevision,
  title = {LLaVA-OneVision: Easy Visual Task Transfer},
  author = {Bo Li and Yuanhan Zhang and Dong Guo and Renrui Zhang and Feng Li and Hao Zhang and Kaichen Zhang and Peiyuan Zhang and Yanwei Li and Ziwei Liu and Chunyuan Li},
  journal = {arXiv preprint arXiv:2408.03326},
  year = {2024}
}

@misc{li2024llavanext,
    title={LLaVA-NeXT: What Else Influences Visual Instruction Tuning Beyond Data?},
    url={https://llava-vl.github.io/blog/2024-05-25-llava-next-ablations/},
    author={Li, Bo and Zhang, Hao and Zhang, Kaichen and Guo, Dong and Zhang, Yuanhan and Zhang, Renrui and Li, Feng and Liu, Ziwei and Li, Chunyuan},
    month={May},
    year={2024}
}

@misc{chen2024huatuogptvision,
  title={HuatuoGPT-Vision, Towards Injecting Medical Visual Knowledge into Multimodal LLMs at Scale}, 
  author={Junying Chen and Chi Gui and Ruyi Ouyang and Anningzhe Gao and Shunian Chen and Guiming Hardy Chen and Xidong Wang and Ruifei Zhang and Zhenyang Cai and Ke Ji and Guangjun Yu and Xiang Wan and Benyou Wang},
  year={2024},
  eprint={2406.19280},
  archivePrefix={arXiv},
  primaryClass={cs.CV},
  url={https://arxiv.org/abs/2406.19280}, 
}

@misc{lasa2025lingshu,
  title={Lingshu: A Generalist Foundation Model for Unified Multimodal Medical Understanding and Reasoning}, 
  author={LASA Team and Weiwen Xu and Hou Pong Chan and Long Li and Mahani Aljunied and Ruifeng Yuan and Jianyu Wang and Chenghao Xiao and Guizhen Chen and Chaoqun Liu and Zhaodonghui Li and Yu Sun and Junao Shen and Chaojun Wang and Jie Tan and Deli Zhao and Tingyang Xu and Hao Zhang and Yu Rong},
  year={2025},
  eprint={2506.07044},
  archivePrefix={arXiv},
  primaryClass={cs.CL},
  url={https://arxiv.org/abs/2506.07044}, 
}

@misc{google2026medgemma15,
  author = {{Google}},
  title = {google/medgemma-1.5-4b-it},
  year = {2026},
  url = {https://huggingface.co/google/medgemma-1.5-4b-it},
  note = {Official Hugging Face model card}
}

@misc{google2025medgemma27,
  title={MedGemma Technical Report}, 
  author={Andrew Sellergren and Sahar Kazemzadeh and Tiam Jaroensri and Atilla Kiraly and Madeleine Traverse and Timo Kohlberger and Shawn Xu and Fayaz Jamil and Cían Hughes and Charles Lau and Justin Chen and Fereshteh Mahvar and Liron Yatziv and Tiffany Chen and Bram Sterling and Stefanie Anna Baby and Susanna Maria Baby and Jeremy Lai and Samuel Schmidgall and Lu Yang and Kejia Chen and Per Bjornsson and Shashir Reddy and Ryan Brush and Kenneth Philbrick and Mercy Asiedu and Ines Mezerreg and Howard Hu and Howard Yang and Richa Tiwari and Sunny Jansen and Preeti Singh and Yun Liu and Shekoofeh Azizi and Aishwarya Kamath and Johan Ferret and Shreya Pathak and Nino Vieillard and Ramona Merhej and Sarah Perrin and Tatiana Matejovicova and Alexandre Ramé and Morgane Riviere and Louis Rouillard and Thomas Mesnard and Geoffrey Cideron and Jean-bastien Grill and Sabela Ramos and Edouard Yvinec and Michelle Casbon and Elena Buchatskaya and Jean-Baptiste Alayrac and Dmitry Lepikhin and Vlad Feinberg and Sebastian Borgeaud and Alek Andreev and Cassidy Hardin and Robert Dadashi and Léonard Hussenot and Armand Joulin and Olivier Bachem and Yossi Matias and Katherine Chou and Avinatan Hassidim and Kavi Goel and Clement Farabet and Joelle Barral and Tris Warkentin and Jonathon Shlens and David Fleet and Victor Cotruta and Omar Sanseviero and Gus Martins and Phoebe Kirk and Anand Rao and Shravya Shetty and David F. Steiner and Can Kirmizibayrak and Rory Pilgrim and Daniel Golden and Lin Yang},
  year={2025},
  eprint={2507.05201},
  archivePrefix={arXiv},
  primaryClass={cs.AI},
  url={https://arxiv.org/abs/2507.05201}, 
}

\clearpage
\newpage
\appendix

\section{AI Usage Declaration}
We declare that the textual content of this paper was originally drafted by the authors. We utilized Gemini-3-Pro solely for proofreading and grammatical error correction. In addition, some illustrative figures were generated with assistance from Nano-Banana. All authors have carefully reviewed the polished text and figures, and take full responsibility for the final content.

\section{Data Availability}
Because \textsc{Dental-TriageBench} is built from real clinical cases, the dataset cannot be released for unrestricted public access. To support responsible research use, we plan to release the benchmark under a controlled access mechanism: researchers will be able to obtain the data through an application process subject to institutional approval and data use conditions.

\section{Ethics Review and Data Governance}
Medical records were retrospectively retrieved from a university-affiliated dental teaching hospital, with all patient data anonymized prior to annotation and analysis. The study protocol was approved by the relevant institutional ethics review board.

\section{Human Rater Recruitment and Compensation}
The three junior dentists who participated in our human baseline evaluation were recruited through voluntary participation. All raters were licensed junior dentists with relevant clinical training, and none were recruited through crowdsourcing platforms. Each rater was compensated for their time and professional effort at a rate aligned with reasonable local professional wages, which we considered appropriate for the participants' expertise and local compensation standards.

\section{Triage Label Taxonomy}
\label{sec:appendix_triage_taxonomy}
\begin{figure}[htbp!]
  \centering
  \includegraphics[width=0.8\linewidth]{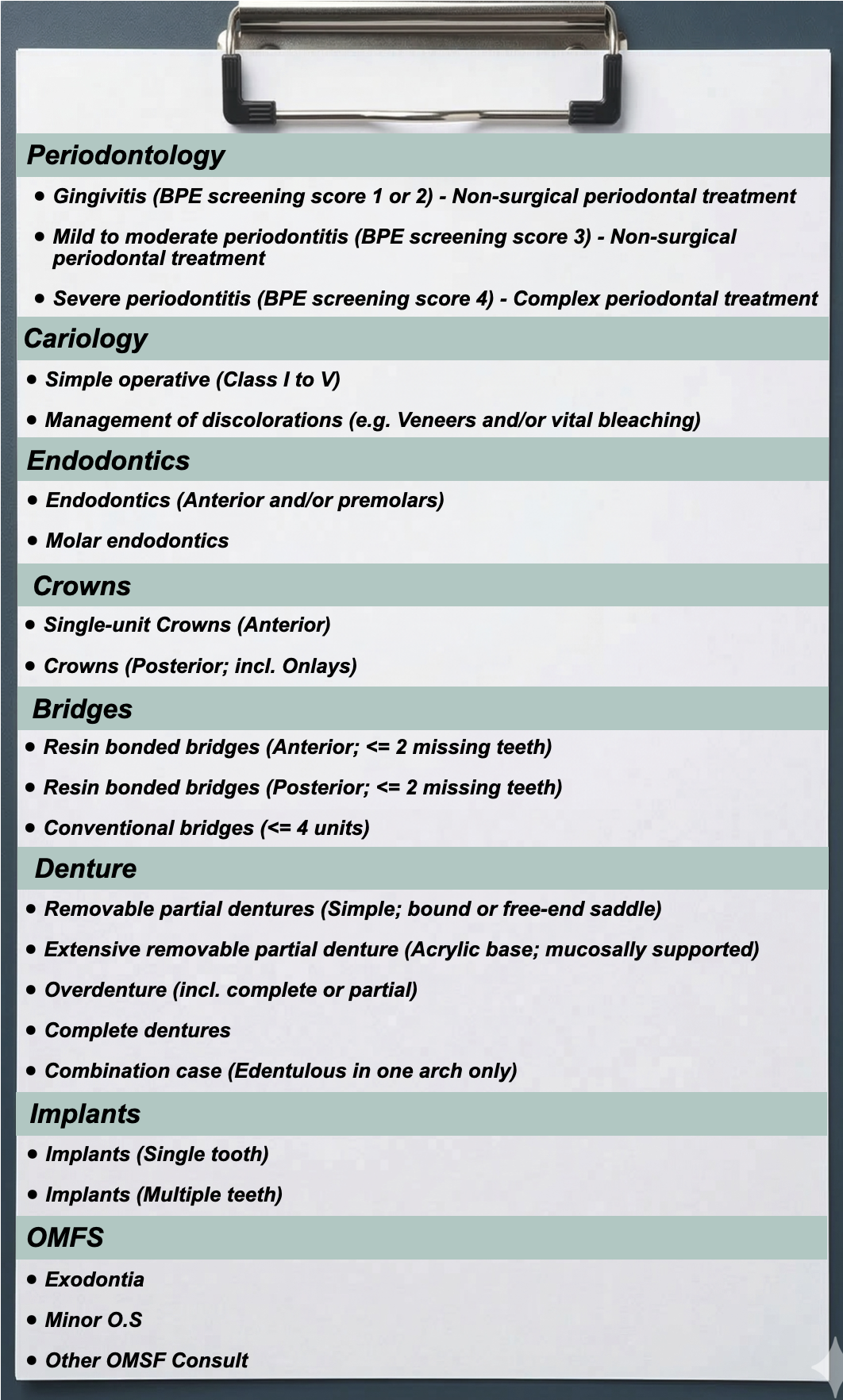}
  \caption{\textbf{Hierarchical taxonomy of the dental triage label space in our benchmark.} The taxonomy is organized into two levels: 8 coarse-grained triage domains (e.g., Periodontology, Endodontics, and OMFS) and 22 fine-grained triage categories, which specify treatment-oriented referral categories within each coarse-grained triage domain.}
  \label{fig:label_hierachy}
\end{figure}

our benchmark adopts a \textbf{hierarchical triage taxonomy} adapted from authentic hospital outpatient triage forms and further refined by senior dental experts. The complete taxonomy is illustrated in Figure~\ref{fig:label_hierachy}. The taxonomy is defined at two levels:
\begin{itemize}
  \item \textbf{Coarse-grained triage domains (8 categories): domain-level routing.}  
  The 8 coarse-grained triage domains (e.g., Periodontology, Endodontics, and OMFS) correspond to the major clinical departments involved in dental triage. This level answers the high-level question: \textit{Which specialty should evaluate or manage this patient?} In our further analyses, these coarse-grained triage domains are used to study referral breadth, case complexity, and omission risk.

  \item \textbf{Fine-grained triage labels (22 categories): treatment-oriented complexity stratification.}  
  Within each coarse-grained triage domain, the 22 fine-grained triage labels capture clinically meaningful distinctions in referral type and treatment complexity. In academic dental hospitals, such distinctions are essential for assigning patients to the appropriate level of provider (e.g., undergraduate students versus postgraduate specialists) and for allocating suitable treatment time and resources. The primary benchmark prediction task is therefore defined at this fine-grained level.
\end{itemize}

For instance, within the Within each coarse-grained triage domain of \textit{Endodontics}, our taxonomy distinguishes between the fine-grained triage labels \textit{Endodontics (Anterior/Premolars)} and \textit{Molar Endodontics}. From a purely symptomatic perspective, both may require root canal treatment. However, from a triage and resource allocation perspective, an anterior tooth often involves a simpler canal system that can be handled by a junior practitioner within a standard appointment slot, whereas a molar typically presents a more complex canal anatomy, requiring a more experienced endodontist, specialized equipment such as an operating microscope, and substantially more chair time.

\paragraph{Distinguishing Triage from Diagnosis}
It is important to clarify that the 22 fine-grained triage labels are \textbf{not definitive pathological diagnoses}. In dentistry, definitive diagnosis (e.g., irreversible pulpitis) often requires additional clinical examinations, such as cold testing or electric pulp testing, which cannot be performed from an OPG alone. Instead, these labels represent \textbf{treatment-oriented triage categories} that support downstream referral and resource planning. This formulation keeps the benchmark aligned with the real clinical scope of triage: making the safest and most efficient referral decision from limited multimodal evidence (OPG + chief complaint), rather than forcing the model to infer a definitive diagnosis beyond the available information.

\section{Output Parsing and Evaluation Metrics}
\label{sec:appendix_output_and_metrics}
\subsection{Output Parsing and Structured Output Extraction}
\label{subsec:appendix_output}
To ensure consistent evaluation across heterogeneous MLLMs, we standardized all model outputs into a unified JSON schema before computing metrics. Each evaluated model was instructed to directly generate its prediction in a predefined JSON format containing the structured triage decision fields required for evaluation. The target schema specifies the full set of triage keys and enables deterministic extraction of fine-grained labels and, when available, the accompanying rationale.

Because some models occasionally produced malformed JSON or omitted required keys, we adopted a two-stage parsing pipeline. First, each model was given up to three attempts to produce a valid JSON output under the same task instruction, with retries triggered only when the output could not be parsed as valid JSON or failed schema validation due to missing required fields. If the model still failed after three attempts, we applied an external reformatting step using Gemini-Flash-2.0~\cite{google2025gemini2flash}. Importantly, this step was used only to convert the model's original response into the target schema and did not perform a new clinical prediction.

The reformatting model received the raw model output and was instructed to preserve the original semantic content while reorganizing it into the predefined JSON structure. It was not provided with ground-truth labels and was not allowed to infer additional findings, add unsupported referrals, or revise the clinical meaning of the original answer. In other words, this fallback mechanism was strictly schema-preserving rather than prediction-correcting. The prompt used for this reformatting step is provided in Table~\ref{fig:output_format}.

After structured parsing, the predicted fine-grained triage labels were deterministically mapped into binary label vectors for evaluation. Coarse-grained domain predictions were then obtained by aggregating the corresponding fine-grained labels according to the benchmark taxonomy.

\subsection{Evaluation Metrics}
\label{subsec:appendix_metrics}
We evaluate structured triage prediction under both the \textbf{fine-grained 22-label space} and the \textbf{coarse-grained 8-domain space}. In both settings, each case is treated as a multi-label prediction problem. Let $\mathcal{L}$ denote the label set, where $|\mathcal{L}|=22$ for the fine-grained task and $|\mathcal{L}|=8$ for the coarse-grained task. For each case $i \in \{1,\dots,N\}$, the ground-truth and predicted label vectors are denoted by $\mathbf{y}_i, \hat{\mathbf{y}}_i \in \{0,1\}^{|\mathcal{L}|}$.

We report four complementary metrics: \textbf{Macro-F1}, \textbf{Micro-F1}, \textbf{Macro-Recall}, and \textbf{Exact Match Ratio (EMR)}. These metrics together capture class-balanced discrimination, overall label recovery, sensitivity to omission, and strict case-level correctness under imbalanced multi-label prediction.

\paragraph{Macro-F1}
For each label $\ell \in \mathcal{L}$, we compute its precision $P_\ell$, recall $R_\ell$, and F1 score:
\[
P_\ell = \frac{TP_\ell}{TP_\ell + FP_\ell}, \qquad
R_\ell = \frac{TP_\ell}{TP_\ell + FN_\ell},
\]
\[
F1_\ell = \frac{2 P_\ell R_\ell}{P_\ell + R_\ell}.
\]
Macro-F1 is then defined as the unweighted average across labels:
\[
\text{Macro-F1} = \frac{1}{|\mathcal{L}|}\sum_{\ell \in \mathcal{L}} F1_\ell.
\]
Macro-F1 measures how well a model performs \emph{across all referral categories equally}, regardless of label frequency. In our setting, this is important because \textsc{Dental-TriageBench} exhibits a pronounced long-tail distribution, and clinically less frequent referral categories (e.g., Bridges or Implants) remain important despite having fewer samples.

\paragraph{Micro-F1}
Micro-F1 aggregates true positives, false positives, and false negatives over all labels before computing the F1 score:
\[
P_{\text{micro}} = \frac{\sum_{\ell} TP_\ell}{\sum_{\ell} (TP_\ell + FP_\ell)},
\]
\[
R_{\text{micro}} = \frac{\sum_{\ell} TP_\ell}{\sum_{\ell} (TP_\ell + FN_\ell)},
\]
\[
\text{Micro-F1} = \frac{2 P_{\text{micro}} R_{\text{micro}}}{P_{\text{micro}} + R_{\text{micro}}}.
\]
Micro-F1 reflects \emph{overall label-level prediction quality} across the benchmark, giving more weight to frequent categories. In the dental triage setting, it captures how reliably a model recovers the dominant referral needs over all patient--label decisions, and is therefore useful as a population-level summary of structured triage performance.

\paragraph{Macro-Recall}
Macro-Recall is defined as the unweighted average of per-label recall:
\[
\text{Macro-Recall} = \frac{1}{|\mathcal{L}|}\sum_{\ell \in \mathcal{L}} R_\ell.
\]
We report Macro-Recall separately because recall is clinically especially important in triage. Missing a required referral leaves clinically relevant needs uncovered and may delay downstream care, whereas modest over-referral is often less harmful. Macro-Recall therefore directly reflects a model's ability to \emph{cover the full range of necessary domain referrals}, including less frequent categories.

\paragraph{Exact Match Ratio (EMR)}
Exact Match Ratio measures the proportion of cases for which the predicted label set exactly matches the ground-truth label set:
\[
\text{EMR} = \frac{1}{N}\sum_{i=1}^{N}\mathbb{I}(\hat{\mathbf{y}}_i = \mathbf{y}_i),
\]
where $\mathbb{I}(\cdot)$ is the indicator function. EMR is a strict case-level metric: a prediction is counted as correct only if \emph{all} required labels are recovered and no extra labels are added. In our task, this metric is clinically meaningful because dental triage is inherently multi-label and incomplete referral sets can be unsafe even when the main domain is correctly identified. EMR therefore captures whether a model can produce a \emph{fully correct and complete triage decision} for an individual patient.

Taken together, these four metrics provide complementary views of performance. Macro-F1 and Macro-Recall emphasize balanced performance across the full triage taxonomy, including rare but clinically meaningful categories; Micro-F1 summarizes overall label-level accuracy across the dataset; and EMR evaluates whether the model can recover the complete referral set for each case without omission or spurious additions. Reporting all four is therefore important for assessing both \emph{predictive quality} and \emph{clinical safety} in hierarchical multi-label dental triage.

\begin{figure*}[t!]
  \centering
  \includegraphics[width=1.0\textwidth]{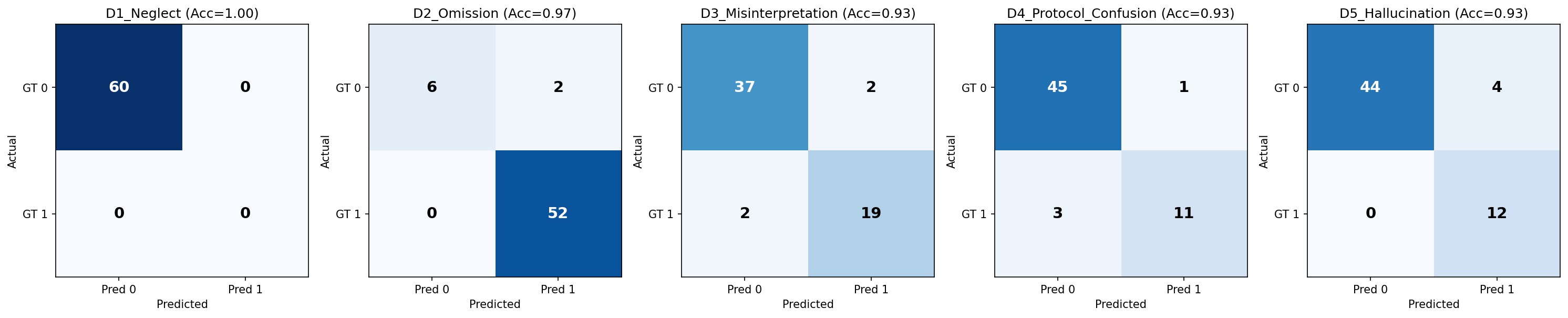}
  \caption{Confusion matrices for human--judge agreement on five failure dimensions over 60 sampled model outputs.}
  \label{fig:confusion_matrices}
\end{figure*}

\begin{table*}[t!]
\centering
\small
\definecolor{lightbluecell}{RGB}{200,225,245}
\definecolor{darkerbluecell}{RGB}{175,205,235}
\definecolor{HeatMapBlue}{RGB}{60,135,215} 

\begin{tabular}{l|cccccccc}
\toprule[1.5pt]
Models & \textbf{Perio} & \textbf{Cariology} & \textbf{Endo} & \textbf{Crowns} & \textbf{Bridges} & \textbf{Denture} & \textbf{Implants} & \textbf{OMFS} \\ \hline
\multicolumn{9}{l}{\cellcolor{lightbluecell}\textit{\textbf{Proprietary MLLMs}}}                    \\ 
Claude-Sonnet-4-5-20250929       & \cellcolor{HeatMapBlue!60}0.598 & \cellcolor{HeatMapBlue!34}0.340 & \cellcolor{HeatMapBlue!45}\underline{0.446} & \cellcolor{HeatMapBlue!37}0.368 & \cellcolor{HeatMapBlue!13}0.130 & \cellcolor{HeatMapBlue!57}0.571 & \cellcolor{HeatMapBlue!26}\underline{0.259} & \cellcolor{HeatMapBlue!82}0.821 \\
Gemini-3-Flash & \cellcolor{HeatMapBlue!65}\underline{0.652} & \cellcolor{HeatMapBlue!44}\underline{0.443} & \cellcolor{HeatMapBlue!34}0.337 & \cellcolor{HeatMapBlue!38}\underline{0.375} & \cellcolor{HeatMapBlue!22}\underline{0.222} & \cellcolor{HeatMapBlue!72}\textbf{0.723} & \cellcolor{HeatMapBlue!24}0.237 & \cellcolor{HeatMapBlue!91}\textbf{0.913} \\
Gemini-3-Pro             & \cellcolor{HeatMapBlue!74}\textbf{0.735} & \cellcolor{HeatMapBlue!37}0.368 & \cellcolor{HeatMapBlue!44}0.444 & \cellcolor{HeatMapBlue!41}\textbf{0.413} & \cellcolor{HeatMapBlue!24}\textbf{0.235} & \cellcolor{HeatMapBlue!61}\underline{0.610} & \cellcolor{HeatMapBlue!33}\textbf{0.333} & \cellcolor{HeatMapBlue!81}0.807 \\
GPT-5.2           & \cellcolor{HeatMapBlue!49}0.494 & \cellcolor{HeatMapBlue!38}0.378 & \cellcolor{HeatMapBlue!34}0.338 & \cellcolor{HeatMapBlue!26}0.255 & \cellcolor{HeatMapBlue!7}0.069 & \cellcolor{HeatMapBlue!57}0.571 & \cellcolor{HeatMapBlue!23}0.226 & \cellcolor{HeatMapBlue!85}\underline{0.850} \\
Grok-4                            & \cellcolor{HeatMapBlue!56}0.561 & \cellcolor{HeatMapBlue!45}\textbf{0.452} & \cellcolor{HeatMapBlue!35}0.354 & \cellcolor{HeatMapBlue!25}0.250 & \cellcolor{HeatMapBlue!13}0.125 & \cellcolor{HeatMapBlue!58}0.577 & \cellcolor{HeatMapBlue!12}0.120 & \cellcolor{HeatMapBlue!70}0.695 \\
Kimi-K2.5                         & \cellcolor{HeatMapBlue!44}0.435 & \cellcolor{HeatMapBlue!44}0.436 & \cellcolor{HeatMapBlue!40}0.396 & \cellcolor{HeatMapBlue!25}0.245 & \cellcolor{HeatMapBlue!15}0.148 & \cellcolor{HeatMapBlue!48}0.482 & \cellcolor{HeatMapBlue!21}0.207 & \cellcolor{HeatMapBlue!80}0.797 \\
Qwen-VL-Max-Latest                & \cellcolor{HeatMapBlue!45}0.449 & \cellcolor{HeatMapBlue!41}0.410 & \cellcolor{HeatMapBlue!45}\textbf{0.453} & \cellcolor{HeatMapBlue!31}0.311 & \cellcolor{HeatMapBlue!15}0.154 & \cellcolor{HeatMapBlue!47}0.469 & \cellcolor{HeatMapBlue!20}0.203 & \cellcolor{HeatMapBlue!73}0.734 \\ 
\multicolumn{9}{l}{\cellcolor{lightbluecell}\textit{\textbf{Open-source MLLMs}}}                     \\ 
GLM-4.1V-9B-Thinking               & \cellcolor{HeatMapBlue!13}0.133 & \cellcolor{HeatMapBlue!4}0.039 & \cellcolor{HeatMapBlue!35}\underline{0.349} & \cellcolor{HeatMapBlue!5}0.053 & \cellcolor{HeatMapBlue!0}0.000 & \cellcolor{HeatMapBlue!15}0.154 & \cellcolor{HeatMapBlue!8}0.080 & \cellcolor{HeatMapBlue!26}0.259 \\
InternVL3\_5-8B                    & \cellcolor{HeatMapBlue!26}0.258 & \cellcolor{HeatMapBlue!22}0.218 & \cellcolor{HeatMapBlue!20}0.197 & \cellcolor{HeatMapBlue!7}0.065 & \cellcolor{HeatMapBlue!8}\textbf{0.080} & \cellcolor{HeatMapBlue!31}0.310 & \cellcolor{HeatMapBlue!14}0.136 & \cellcolor{HeatMapBlue!14}0.143 \\
MiMo-VL-7B-RL                      & \cellcolor{HeatMapBlue!29}0.288 & \cellcolor{HeatMapBlue!34}\underline{0.338} & \cellcolor{HeatMapBlue!30}0.297 & \cellcolor{HeatMapBlue!11}0.114 & \cellcolor{HeatMapBlue!7}0.074 & \cellcolor{HeatMapBlue!33}\underline{0.329} & \cellcolor{HeatMapBlue!8}0.078 & \cellcolor{HeatMapBlue!61}\underline{0.612} \\
Ministral-3-14B-Instruct-2512      & \cellcolor{HeatMapBlue!38}\textbf{0.376} & \cellcolor{HeatMapBlue!32}0.319 & \cellcolor{HeatMapBlue!39}\textbf{0.385} & \cellcolor{HeatMapBlue!20}\textbf{0.195} & \cellcolor{HeatMapBlue!7}0.067 & \cellcolor{HeatMapBlue!42}\textbf{0.420} & \cellcolor{HeatMapBlue!18}\underline{0.179} & \cellcolor{HeatMapBlue!68}\textbf{0.679} \\
Qwen2.5-VL-7B-Instruct             & \cellcolor{HeatMapBlue!27}0.274 & \cellcolor{HeatMapBlue!4}0.039 & \cellcolor{HeatMapBlue!32}0.320 & \cellcolor{HeatMapBlue!10}0.098 & \cellcolor{HeatMapBlue!8}\underline{0.077} & \cellcolor{HeatMapBlue!15}0.154 & \cellcolor{HeatMapBlue!8}0.078 & \cellcolor{HeatMapBlue!49}0.489 \\
Qwen3-VL-8B-Instruct               & \cellcolor{HeatMapBlue!16}0.161 & \cellcolor{HeatMapBlue!36}\textbf{0.358} & \cellcolor{HeatMapBlue!14}0.143 & \cellcolor{HeatMapBlue!0}0.000 & \cellcolor{HeatMapBlue!8}\textbf{0.080} & \cellcolor{HeatMapBlue!29}0.286 & \cellcolor{HeatMapBlue!19}\textbf{0.185} & \cellcolor{HeatMapBlue!45}0.448 \\
LLaVA-onevision-qwen2-7b     & \cellcolor{HeatMapBlue!31}\underline{0.306} & \cellcolor{HeatMapBlue!22}0.215 & \cellcolor{HeatMapBlue!11}0.111 & \cellcolor{HeatMapBlue!14}\underline{0.140} & \cellcolor{HeatMapBlue!0}0.000 & \cellcolor{HeatMapBlue!0}0.000 & \cellcolor{HeatMapBlue!12}0.120 & \cellcolor{HeatMapBlue!18}0.181 \\
LLaVA-v1.6-mistral-7b-hf           & \cellcolor{HeatMapBlue!17}0.169 & \cellcolor{HeatMapBlue!4}0.044 & \cellcolor{HeatMapBlue!0}0.000 & \cellcolor{HeatMapBlue!0}0.000 & \cellcolor{HeatMapBlue!0}0.000 & \cellcolor{HeatMapBlue!0}0.000 & \cellcolor{HeatMapBlue!4}0.044 & \cellcolor{HeatMapBlue!0}0.000 \\ 
\multicolumn{9}{l}{\cellcolor{lightbluecell}\textit{\textbf{Medical MLLMs}}}                       \\ 
HuatuoGPT-Vision-7B                & \cellcolor{HeatMapBlue!32}\textbf{0.317} & \cellcolor{HeatMapBlue!7}\underline{0.073} & \cellcolor{HeatMapBlue!7}0.070 & \cellcolor{HeatMapBlue!6}0.056 & \cellcolor{HeatMapBlue!0}0.000 & \cellcolor{HeatMapBlue!22}\underline{0.222} & \cellcolor{HeatMapBlue!4}0.039 & \cellcolor{HeatMapBlue!19}0.191 \\
Lingshu-7B                         & \cellcolor{HeatMapBlue!13}0.133 & \cellcolor{HeatMapBlue!4}0.043 & \cellcolor{HeatMapBlue!14}0.136 & \cellcolor{HeatMapBlue!15}\underline{0.146} & \cellcolor{HeatMapBlue!8}\textbf{0.080} & \cellcolor{HeatMapBlue!24}\textbf{0.235} & \cellcolor{HeatMapBlue!8}0.082 & \cellcolor{HeatMapBlue!52}\underline{0.523} \\
MedGemma-1.5-4b-it                 & \cellcolor{HeatMapBlue!10}0.104 & \cellcolor{HeatMapBlue!7}0.070 & \cellcolor{HeatMapBlue!18}\underline{0.179} & \cellcolor{HeatMapBlue!17}\textbf{0.167} & \cellcolor{HeatMapBlue!0}0.000 & \cellcolor{HeatMapBlue!7}0.065 & \cellcolor{HeatMapBlue!14}\textbf{0.143} & \cellcolor{HeatMapBlue!21}0.205 \\
MedGemma-27b-it                    & \cellcolor{HeatMapBlue!27}\underline{0.265} & \cellcolor{HeatMapBlue!36}\textbf{0.361} & \cellcolor{HeatMapBlue!42}\textbf{0.421} & \cellcolor{HeatMapBlue!14}0.143 & \cellcolor{HeatMapBlue!7}\underline{0.071} & \cellcolor{HeatMapBlue!21}0.209 & \cellcolor{HeatMapBlue!12}\underline{0.122} & \cellcolor{HeatMapBlue!54}\textbf{0.537} \\
\bottomrule[1.5pt]
\end{tabular}
\caption{Class-wise F1 scores on the 8 coarse-grained domains. Within each model domain, the best results are in \textbf{bold} and the second-best results are \underline{underlined}.}
\label{tab:class_wise_F1_tab}
\end{table*}

\section{Additional Results}
\subsection{Individual Human Baseline Results}
\label{subsec:appendix_human_baseline}
To ensure a fair comparison between human raters and MLLMs, we provided the three junior dentists with the same task instructions used for model evaluation. The full instruction prompt is shown in Table~\ref{fig:task_instruction}.

To complement the averaged human baseline reported in the main paper, we provide the individual performance of the three junior dentists in Table~\ref{tab:human_individual_results}. Although minor variation exists across raters, their overall performance is relatively consistent at both the fine-grained and coarse-grained levels. In particular, all three dentists substantially outperform current MLLMs on the fine-grained triage task, while showing similar strength on coarse-grained specialty routing. These results suggest that the human baseline is not driven by a single unusually strong or weak annotator. We therefore report the average across the three junior dentists in the main paper as a more stable and representative estimate of human performance.

\begin{table*}[t!]
\centering
\footnotesize
\setlength{\tabcolsep}{3.5pt}
\resizebox{\textwidth}{!}{%
\begin{tabular}{l|cccc|cccc}
\toprule[1.5pt]
\multirow{2}{*}{\textbf{Rater}} & \multicolumn{4}{c|}{\textbf{Fine-grained (22 labels)}} & \multicolumn{4}{c}{\textbf{Coarse-grained (8 specialties)}} \\
\cmidrule(lr){2-5}\cmidrule(lr){6-9}
& \textbf{Macro-F1} & \textbf{Macro-Rec} & \textbf{Micro-F1} & \textbf{Exact-Match}
& \textbf{Macro-F1} & \textbf{Macro-Rec} & \textbf{Micro-F1} & \textbf{Exact-Match} \\
\midrule
Junior Dentist 1 & 0.394 & 0.440 & 0.511 & 0.049 & 0.567 & 0.647 & 0.696 & 0.203 \\
Junior Dentist 2 & 0.409 & 0.439 & 0.520 & 0.077 & 0.547 & 0.604 & 0.730 & 0.260 \\
Junior Dentist 3 & 0.402 & 0.413 & 0.546 & 0.122 & 0.571 & 0.586 & 0.754 & 0.354 \\
\midrule
\textbf{Average} & \textbf{0.402} & \textbf{0.431} & \textbf{0.525} & \textbf{0.083}
& \textbf{0.562} & \textbf{0.613} & \textbf{0.727} & \textbf{0.272} \\
\bottomrule[1.5pt]
\end{tabular}%
}
\caption{Individual performance of the three junior dentists on \textsc{Dental-TriageBench}. We report the same evaluation metrics as in the main paper.}
\label{tab:human_individual_results}
\end{table*}

\subsection{Class-wise performance across 8 coarse-grained domains}
\label{subsec:appendix_class_wise_F1}
Table~\ref{tab:class_wise_F1_tab} reports class-wise F1 scores on the 8 coarse-grained domains. A consistent pattern across model families is that performance is much stronger on frequent and visually salient categories such as \textit{OMFS} and \textit{Perio}, but substantially weaker on lower-support categories such as \textit{Bridges}, \textit{Implants}, and \textit{Crowns}. This class-wise disparity reflects the pronounced label imbalance of \textsc{Dental-TriageBench} and provides additional context for the gap between \textit{Micro-F1} and \textit{Macro-F1} in Table~\ref{tab:merged_fine_coarse_tab}: micro-averaging is driven more by high-frequency categories, while macro-averaging weights rare classes equally and is therefore more sensitive to failures on the long tail.

\section{Failure Taxonomy and LLM-as-Judge Details}
\subsection{Definitions of the Five Failure Dimensions}
\label{subsec:appendix_dimension_definition}

To move beyond aggregate set-prediction metrics, we characterize model errors using five clinician-grounded failure dimensions. These dimensions are designed to identify the \emph{root cause} of a model's failure by comparing the model-generated rationale against the expert-annotated ground-truth rationale and the original patient complaint. Importantly, the taxonomy is applied primarily at the level of \textbf{reasoning}, rather than inferred solely from final label discrepancies. A model may therefore exhibit a failure dimension even when its final predicted triage labels are partially or fully correct, if its rationale reveals flawed clinical justification.

The five dimensions are treated as \textbf{strictly mutually exclusive at the level of each specific clinical finding}. In other words, for any individual symptom, lesion, or anatomical issue, we assign only one underlying failure type according to a strict hierarchy. The attribution proceeds in the following order: 
(1) \textbf{Hallucination}, 
(2) \textbf{Chief Complaint Neglect}, 
(3) \textbf{OPG Finding Omission}, 
(4) \textbf{OPG Misinterpretation}, and 
(5) \textbf{Protocol Misalignment}. 
This ordering ensures that downstream errors are attributed to the earliest underlying failure. For example, if the model misjudges the severity of a lesion on OPG, any subsequent missing referral caused by that misjudgment is attributed to \textbf{OPG Misinterpretation} rather than \textbf{Protocol Misalignment}.
We define the five failure dimensions as follows.

\paragraph{D1: Chief Complaint Neglect.}
This dimension captures failures in which the model's rationale completely ignores the core symptom, concern, or treatment request expressed in the patient's chief complaint. It is triggered only when the text complaint contains clinically relevant information that should contribute to triage, but the model fails to acknowledge it at all in its reasoning. This category does not apply when the complaint is mentioned but interpreted imperfectly; rather, it is reserved for complete neglect of the salient textual signal.

\paragraph{D2: OPG Finding Omission.}
This dimension captures failures in which the model completely fails to mention an important radiographic finding that is explicitly identified in the expert ground-truth rationale. Typical examples include missing visible caries, bone loss, impacted teeth, retained roots, or other clinically relevant OPG abnormalities. Crucially, D2 is assigned only when the model omits the finding entirely. If the model does mention the correct finding or anatomical region but misjudges its severity or pathological nature, the error is instead categorized as \textbf{D3: OPG Misinterpretation}.

\paragraph{D3: OPG Misinterpretation.}
This dimension captures failures in which the model identifies the relevant lesion, anatomical region, or radiographic abnormality, but misinterprets its clinical severity, extent, or pathological depth relative to the expert rationale. Examples include describing deep caries as superficial, severe periodontal destruction as mild disease, or a non-restorable tooth as restorable. D3 is intended to capture incorrect visual-clinical grounding rather than total omission. Under the hierarchy, once D3 is assigned for a finding, downstream triage mistakes caused by this initial misjudgment are not separately counted as protocol errors.

\paragraph{D4: Protocol Misalignment.}
This dimension captures failures in which the model correctly understands the relevant pathology or clinical finding, but still maps it to the wrong triage label under the clinical referral rules. D4 is therefore the most restrictive category: it is assigned only when there is no preceding error of complaint neglect, OPG omission, or OPG misinterpretation for that specific finding. In other words, the pathology is perceived correctly, but the final referral decision does not follow the intended clinical triage protocol. Examples include assigning cariology when the rationale correctly describes caries extending to the pulp, or failing to route a correctly identified impacted wisdom tooth to the appropriate OMFS category. The full expert-authored Clinical Triage Protocol can be found at the end of the Table~\ref{fig:llm_as_j}. 

\paragraph{D5: Hallucination.}
\label{subsec:appendix_human_agreenment}
This dimension captures failures in which the model fabricates symptoms, anatomical structures, lesions, or radiographic findings that are unsupported by either the patient complaint or the expert ground-truth rationale. Hallucination is triggered solely from the generated reasoning text and is independent of whether the final triage labels happen to be correct. For example, if the model invents an impacted tooth, a lesion, or a symptom not present in the case, the output is marked as D5 even if the predicted referral set coincidentally overlaps with the ground truth.

\paragraph{Attribution principle.}
The purpose of this taxonomy is not to count every observable downstream mistake independently, but to identify the earliest clinically meaningful source of failure in the model's reasoning process. This design helps distinguish perception failures from decision-rule failures and prevents the same underlying mistake from being counted multiple times across dimensions. As a result, the taxonomy provides a more interpretable account of how multimodal LLMs fail in dental triage. And the detailed prompt of scoring guideline to the LLM Judge is demonstrated in Table~\ref{fig:llm_as_j}.

\subsection{Human--Judge Agreement}
To assess the reliability of our LLM-as-judge failure attribution pipeline, we conducted a human verification study on a randomly sampled subset of benchmark outputs. Specifically, we first randomly sampled 20 cases from \textsc{Dental-TriageBench}, and then for each case randomly selected outputs from 3 models out of 6 representative models that are included for the failure pattern analysis, yielding 60 model--case outputs in total. For each output, one senior dentist independently assigned binary labels on the five clinician-defined failure dimensions, using the same definitions and scoring criteria provided to the judge model. We then compared the judge's predictions against the human annotations.

Overall, the agreement between the judge and the senior dentist was high, with an average accuracy of 95.3\% across the five failure dimensions. The corresponding confusion matrices are shown in Figure~\ref{fig:confusion_matrices}. Among the five dimensions, \textit{Chief Complaint Neglect} achieved perfect agreement (100\%), although this dimension was absent in the sampled subset and therefore reflects agreement on negative cases only. For the remaining dimensions, agreement remained consistently strong: \textit{OPG Finding Omission} reached 96.7\% accuracy, while \textit{OPG Misinterpretation}, \textit{Protocol Misalignment}, and \textit{Hallucination} each achieved 93.3\% accuracy. These results suggest that the proposed judge is sufficiently reliable for large-scale attribution of clinically grounded failure patterns in our benchmark.

\section{Additional Case Studies}
\subsection{A Representative Case on Modality Ablation}
\label{subsec:appendix_modality_ablation}
\begin{figure}[htbp!]
  \centering
  \includegraphics[width=1\linewidth]{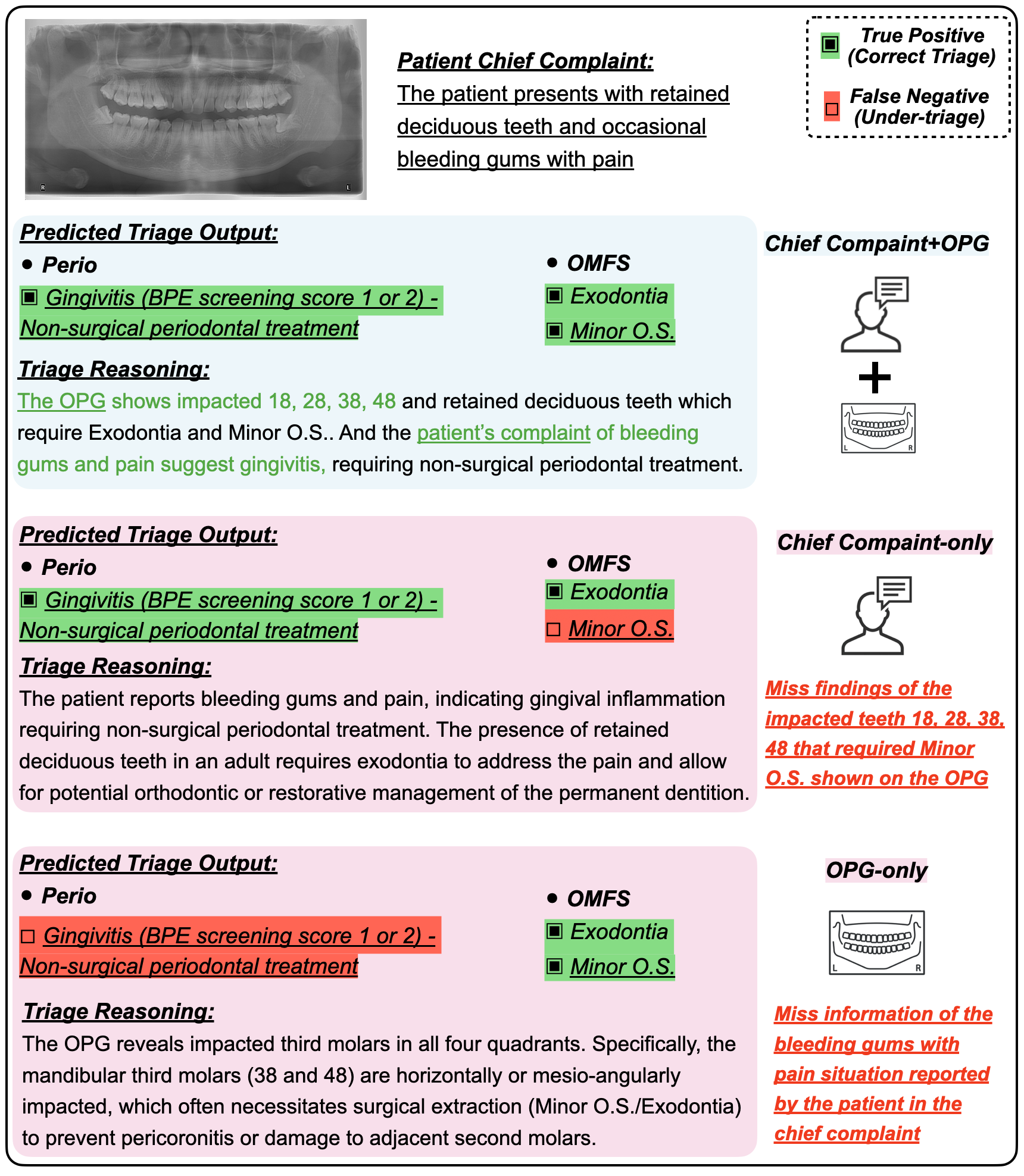}
  \caption{\textbf{A representative modality-ablation case on Gemini-3-Flash.} The chief complaint alone supports \textit{Gingivitis} and \textit{Exodontia}, but misses the surgical complexity required for \textit{Minor O.S.}. With access to the OPG, the model correctly recovers the full OMFS referral plan.}
  \label{fig:modality_ablation_case_study}
\end{figure}

We provide a representative modality-ablation case for Gemini-3-Pro in Figure~\ref{fig:modality_ablation_case_study}. This case illustrates a clear division of labor between the two modalities. The chief complaint, \textit{``retained deciduous teeth and occasional bleeding gums with pain''}, is sufficient to suggest gingival inflammation and some form of extraction need, allowing the text-only model to correctly predict \textit{Gingivitis} and \textit{Exodontia}. However, it does not specify the number, position, or surgical complexity of the affected teeth. As a result, under the complaint-only setting, the model fails to activate \textit{Minor O.S.}, thereby under-calling the full OMFS treatment plan. By contrast, when the OPG is available, the model can directly identify the retained tooth and multiple impacted third molars, which provide the critical visual evidence for surgical complexity and support the correct joint prediction of both \textit{Exodontia} and \textit{Minor O.S.}. This example highlights why dental triage is intrinsically multimodal: text can often capture symptoms and broad treatment intent, whereas radiographic evidence is necessary to determine the full procedural scope and referral complexity.

\subsection{A Failure Case Study on Gemini-3-Pro}
\label{subsec:appendix_failure_case_study}
\begin{figure}[htbp!]
  \centering
  \includegraphics[width=1\linewidth]{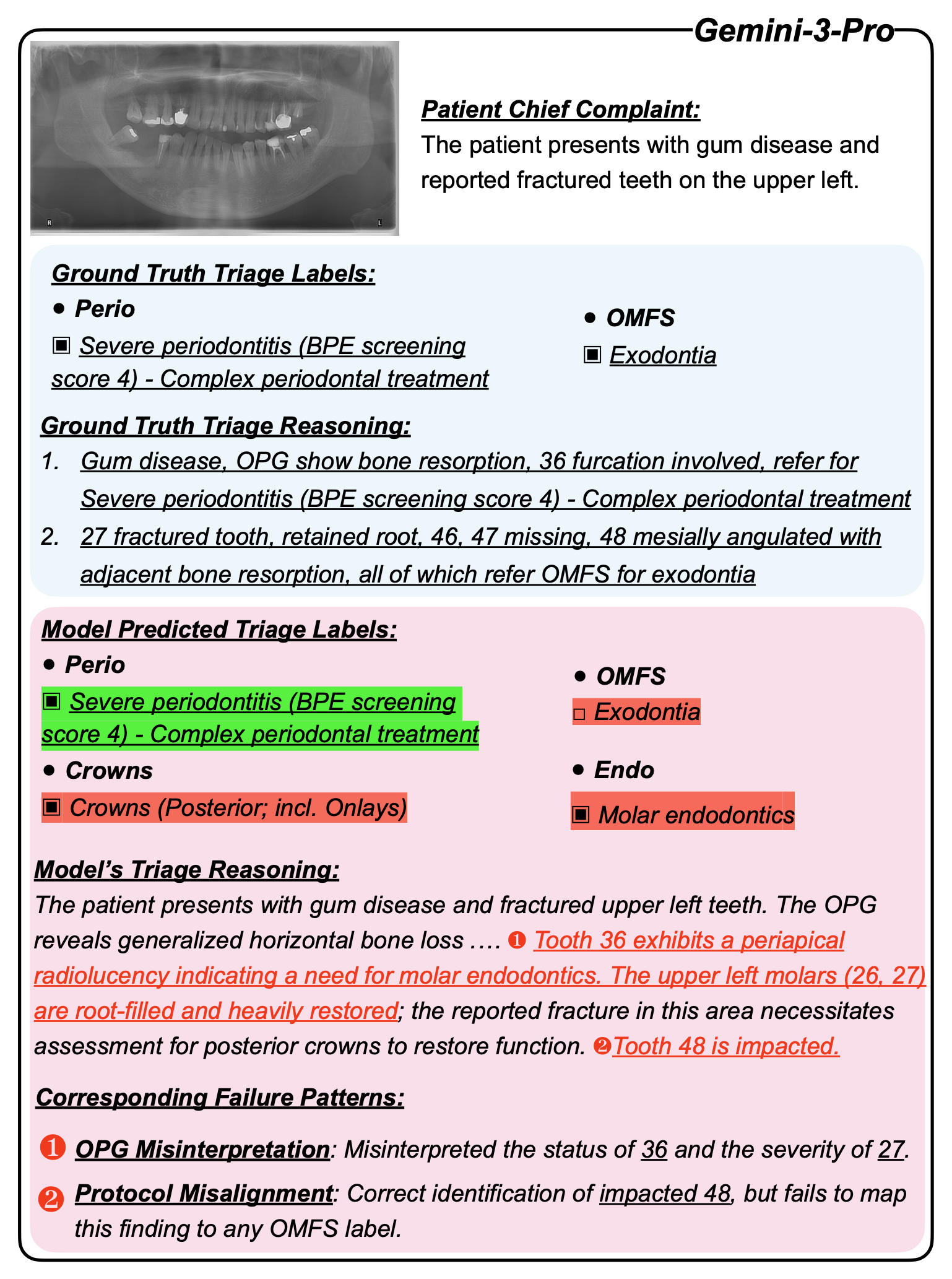}
  \caption{\textbf{Additional failure case study of Gemini-3-Pro.} The model misinterprets the status of tooth 36 and overestimates the severity of tooth 27 on the OPG, leading to incorrect positive predictions for \textit{Molar Endodontics} and \textit{Crowns}. Although it correctly identifies the impaction of tooth 48, it fails to map this finding to the appropriate OMFS referral for \textit{Exodontia}. This example illustrates how OPG misinterpretation can simultaneously produce both over-triage and under-triage errors within the same case.}
  \label{fig:failure_pattern_case_study_gemini}
\end{figure}

We provide an additional failure case study for Gemini-3-Pro in Figure~\ref{fig:failure_pattern_case_study_gemini}. In this case, the model misinterprets both the status of tooth 36 and the severity of tooth 27 on the OPG. These OPG misinterpretation errors lead it to assign incorrect triage labels, namely \textit{Molar Endodontics} and \textit{Crowns}, resulting in false positives and thus over-triage. Although the model correctly recognizes the impaction of tooth 48, it fails to translate this finding into the appropriate OMFS referral for \textit{Exodontia}, leading to under-triage.

\section{Prompt Template Details}
\label{sec: prompt}
\subsection{Prompt Template for Model Inference}
\label{subsec:appendix_inference_prompt}
Table~\ref{fig:task_instruction} illustrates the task instruction to all models for 0-shot inference.
\subsection{Prompt Template for Output Formatting}
\label{subsec:appendix_format_prompt}
We present the output formatting prompt in Table~\ref{fig:output_format} to another LLM  when the model output is inconsistent with the standardized JSON output format
\subsection{Prompt Template for LLM Judge}
\label{subsec:appendix_llm_as_j_prompt}
Table~\ref{fig:llm_as_j} displays the scoring guideline for LLM judge used in the Section~\ref{sec: failure_analysis}.

\onecolumn

\clearpage

\clearpage

\clearpage

\begin{center}
\begin{tcolorbox}[colback=lightbluebg!30!white,colframe=blueframe,breakable,title=Prompt for task instruction for model inference]
\begin{VerbatimWrap}
You are a triage expert and dentist, you are given a patient's chief complaint and one OPG image, and a list of triage/treatment labels, you need to triage the patient to the most appropriate triage/treatment labels and provide your reasoning for the triage (where the label is 1). The reasoning should be a string that explains your triage decision based on the chief complaint and OPG image findings. Please give your response exactly in the following JSON format where 0 stands for no, 1 stands for yes:

```json
{{  
    "triage_output": {{
        "Perio": {{
            "Gingivitis (BPE screening score 1 or 2) - Non-surgical periodontal treatment": 0,
            "Mild to moderate periodontitis (BPE screening score 3) - Non-surgical periodontal treatment": 0,
            "Severe periodontitis (BPE screening score 4) - Complex periodontal treatment": 0
        }},
        "Cariology": {{
            "Simple operative (Class I to V)": 0,
            "Management of discolorations (e.g. Veneers and/or vital bleaching)": 0
        }},
        "Endo": {{
            "Endodontics (Anterior and/or premolars)": 0,
            "Molar endodontics": 0
        }},
        "Crowns": {{
            "Single-unit Crowns (Anterior)": 0,
            "Crowns (Posterior; incl. Onlays)": 0
        }},
        "Bridges": {{
            "Resin bonded bridges (Anterior; <= 2 missing teeth)": 0,
            "Resin bonded bridges (Posterior; <= 2 missing teeth)": 0,
            "Conventional bridges (<= 4 units)": 0
        }},
        "Denture": {{
            "Removable partial dentures (Simple; bound or free-end saddle)": 0,
            "Extensive removable partial denture (Acrylic base; mucosally supported)": 0,
            "Overdenture (incl. complete or partial)": 0,
            "Complete dentures": 0,
            "Combination case (Edentulous in one arch only)": 0
        }},
        "Implants": {{
            "Implants (Single tooth)": 0,
            "Implants (Multiple teeth)": 0
        }},
        “OMFS”: {{
            "Exodontia": 0,
            "Minor O.S.": 0,
            "Other OMSF Consult": 0
        }}
    }},
    "reasoning": "Your concise analysis here (max 200 words)"
}}
```
And pay attention to the following:
1. Other OMFS Consult includes the following items: Tumors, cysts, TMJ, fractures, mucosal lesions.
2. Bear in mind that there could be multiple triage/treatment labels that are relevant to the patient, meaning that the patient has multiple conditions or needs multiple treatments. 
3. Your output should be a json object that is strictly formatted to the above json format, please do not add any other text or comments to the output or add any new keys to the json object.

The chief complaint: {chief_complaint}
And the OPG image is:
\end{VerbatimWrap}
\end{tcolorbox}
\vspace{-2mm}
\captionof{table}{The task insturction prompt for model inference}
\label{fig:task_instruction}
\end{center}
\clearpage

\begin{center}
\begin{tcolorbox}[colback=lightbluebg!30!white,colframe=blueframe,breakable,title=Prompt for task instruction for model inference]
\begin{VerbatimWrap}
You are an expert in JSON formatting, you are given an invalid JSON object or free text from another model, you need to format its answer to a valid JSON object. The format should strictly follow this:

{{
    "triage_output": {{
        "Perio": {{
            "Gingivitis (BPE screening score 1 or 2) - Non-surgical periodontal treatment": 0,
            "Mild to moderate periodontitis (BPE screening score 3) - Non-surgical periodontal treatment": 0,
            "Severe periodontitis (BPE screening score 4) - Complex periodontal treatment": 0
        }},
        "Cariology": {{
            "Simple operative (Class I to V)": 0,
            "Management of discolorations (e.g. Veneers and/or vital bleaching)": 0
        }},
        "Endo": {{
            "Endodontics (Anterior and/or premolars)": 0,
            "Molar endodontics": 0
        }},
        "Crowns": {{
            "Single-unit Crowns (Anterior)": 0,
            "Crowns (Posterior; incl. Onlays)": 0
        }},
        "Bridges": {{
            "Resin bonded bridges (Anterior; <= 2 missing teeth)": 0,
            "Resin bonded bridges (Posterior; <= 2 missing teeth)": 0,
            "Conventional bridges (<= 4 units)": 0
        }},
        "Denture": {{
            "Removable partial dentures (Simple; bound or free-end saddle)": 0,
            "Extensive removable partial denture (Acrylic base; mucosally supported)": 0,
            "Overdenture (incl. complete or partial)": 0,
            "Complete dentures": 0,
            "Combination case (Edentulous in one arch only)": 0
        }},
        "Implants": {{
            "Implants (Single tooth)": 0,
            "Implants (Multiple teeth)": 0
        }},
        “OMFS”: {{
            "Exodontia": 0,
            "Minor O.S.": 0,
            "Other OMSF Consult": 0
        }}
    }},
    "reasoning": "model's analysis (in string format, not dictionary or any other format)"
}}

Pay attention:
1. The orginal model output might not contain all the keys in the json object, for those keys that are not present in the original model output, you should set the value to 0.
2. Any other free text that is not related to the key-value pairs, you should convert them to the reasoning field if that makes sense.
3. The orignial input to the model is: 'You are a triage expert. Analyze the patient's chief complaint, medical history, and OPG image. Output the triage decision strictly in JSON format.'

Now the original model output is: {model_output}
\end{VerbatimWrap}
\end{tcolorbox}
\vspace{-2mm}
\captionof{table}{The prompt for standardizing model output}
\label{fig:output_format}
\end{center}
\clearpage

\begin{center}
\begin{tcolorbox}[colback=lightbluebg!30!white,colframe=blueframe,breakable,title=Prompt for llm as judge for failure analysis]
\begin{VerbatimWrap}
# Role
You are an expert Chief Dentist and a rigorous AI evaluator. Your task is to perform a Failure Analysis on an AI model's performance in a Multimodal Dental Triage task.

# Input Data
1. [Patient Input]: 
   - Chief Complaint: {chief_complaint}

2. [Label Discrepancies] (Crucial for Evaluation):
   - Missing Labels (False Negatives - model failed to predict): {missing_labels}
   - Extra Labels (False Positives - model wrongly predicted): {extra_labels}

3. [Ground Truth (GT)]: 
   - GT Reasoning: {gt_reasoning}

4. [Model]: 
   - Model Reasoning: {model_reasoning}

# Core Evaluation Philosophy (Doctor's Workflow)
**CRITICAL:** You must evaluate the model primarily based on its **[Model Reasoning]** compared to the **[GT Reasoning]**. Do NOT use label discrepancies as the sole trigger for finding errors. 
- Example: If the model fabricates a tooth in its reasoning, it is a Hallucination (D5) EVEN IF it accidentally guessed the correct final triage label.
- Example: If the model misjudges a tooth's severity (D3), any downstream missing restorative labels (like missing a Crown or Implant) are considered a CASCADING consequence of D3. Do NOT tag D4 for the missing restoration.

# Task Instruction: FIND THE ROOT CAUSE
Evaluate the model on the following 5 STRICTLY MUTUALLY EXCLUSIVE error dimensions. 
For each specific clinical issue/tooth mentioned in the GT or Model reasoning, assign it to ONLY ONE underlying root cause based on this **Strict Hierarchical Rule**:

**Hierarchy of Evaluation (Evaluate Top to Bottom for each finding):**
1. **Did the model fabricate something? -> D5.** (e.g., mentions an impacted lower wisdom tooth that isn't there, or invents a symptom).
2. **Did the model completely ignore the core text symptom? -> D1.** 
3. **Did the model completely fail to mention an important OPG finding? -> D2.** (e.g., GT mentions generalized bone loss or caries on tooth 11, but the model reasoning never mentions it).
4. **Did the model mention the correct finding but misjudge its physical severity/nature? -> D3.** (e.g., GT says "caries into pulp", but model says "superficial caries"; GT says "severe periodontitis", but model says "mild").
5. **Did the model perceive EVERYTHING correctly about a lesion (NO D1, D2, or D3 errors), but simply assigned the wrong final Triage label? -> ONLY THEN tag D4.**

# The 5 Error Dimensions Definitions:

**[Text] D1: Chief Complaint Neglect**
- The model's reasoning completely ignores the core symptoms/requests in the [Patient Input]. 

**[OPG] D2: OPG Finding Omission**
- The model completely fails to mention important visual findings explicitly listed in the [GT Reasoning]. 
- *Rule:* If the GT mentions a finding and the model reasoning does not mention it at all, it is D2. If the model mentions it but gets the severity wrong, it is D3, NOT D2.

**[OPG] D3: OPG Misinterpretation**
- The model notices the correct lesion/area but misjudges its severity or pathological depth compared to the GT.
- *Rule:* If D3 occurs, DO NOT tag D4 for any missing downstream labels (like missing an extraction or crown) caused by this initial misjudgment. 

**[Triage] D4: Protocol Misalignment**
- **Strict Prerequisite:** The model's reasoning perfectly matches the GT reasoning regarding the anatomical/pathological condition (No D1, D2, or D3 errors for this specific finding). 
- The error is PURELY a failure to map a correctly understood pathology to the correct Triage label based on clinical rules. Refer to the [Appendix: Clinical Triage SOPs (Reference Only)] for more details.

**[Hallucination] D5: Hallucination**
- The model fabricates anatomical structures, lesions, or symptoms in its reasoning that do not exist in the GT or text. 
- *Rule:* Must be triggered based on the reasoning text, regardless of whether the final triage labels are right or wrong.

# Output Format
Return ONLY a valid JSON object. Do not include markdown formatting.
For the "score", output strictly 1 or 0. 
For the "reason", explicitly and briefly state the reasoning of marking this dimension as 1. If the score is 0 for this dimension, leave reason as "N/A".

{
  "D1_Neglect": {"score": 0, "reason": "Explanation / N/A"},
  "D2_Omission": {"score": 0, "reason": "Explanation / N/A"},
  "D3_Misinterpretation": {"score": 0, "reason": "Explanation / N/A"},
  "D4_Protocol_Confusion": {"score": 0, "reason": "Explanation / N/A"},
  "D5_Hallucination": {"score": 0, "reason": "Explanation / N/A"}
}


# [Appendix: Clinical Triage Protocol]
**1. Perio (Periodontal)**
- IF Bone loss <= 1/3 root -> Mild; 1/3-1/2 -> Moderate; > 1/2 -> Severe.
- IF Calculus present -> Require Perio treatment.
- IF Root furcation involved -> Require Complex periodontal treatment.

**2. Cariology vs. Endo**
- IF Caries extends to pulp -> MUST select Endo (NOT Cariology/Simple Operative).
- IF Tooth is non-restorable -> Refer Exodontia.
- IF Discoloration BUT vital pulp & no caries/Endo -> Consider Veneers/bleaching.
- IF Anterior/Premolar AND Molar need RCT -> Select both Endo categories.

**3. Restorations (Crowns / Bridges / Implants)**
- IF Posterior tooth needs RCT -> MUST select Crowns (Posterior/Onlays).
- IF Anterior tooth needs RCT -> NO Crown usually (unless gross structure missing or necrosis discoloration).
- IF Non-wisdom tooth extracted -> MUST consider replacement (Bridges, Dentures, or Implants).

**4. OMFS (Oral Surgery) Routing Rules**
- Exodontia (Simple): No incision/bone removal. E.g., Maxillary vertical wisdom teeth; Erupted mandibular wisdom without adjacent 2nd molar blocking it; Retained roots.
- Minor O.S. (Complex): Requires incision/flap/bone removal. E.g., ALL impacted wisdom teeth; Non-impacted lower wisdom IF adjacent 2nd molar is present.
- Rule Exception: IF simultaneous impacted lower (Minor O.S.) + non-impacted upper (Exodontia) -> Can select Minor O.S. only (done in one session), OR both.
- Other OMFS Consult: Tumors, cysts, TMJ, fractures, mucosal lesions.
\end{VerbatimWrap}
\end{tcolorbox}
\vspace{-2mm}
\captionof{table}{The prompt for LLM-as-Judge for failure analysis}
\label{fig:llm_as_j}
\end{center}

\end{document}